\documentclass[utf8]{SCNSarXiv}  

\usepackage{url,hyperref,microtype,subcaption}

\usepackage{lineno}  
\usepackage[onehalfspacing]{setspace}

\usepackage{doi}
\usepackage[normalem]{ulem}  

\usepackage{graphicx}
\usepackage{booktabs}       
\usepackage{nicefrac}       
\usepackage{microtype}
\usepackage{amssymb}
\usepackage{amsmath}
\usepackage{amsthm}
\usepackage{amsfonts}
\usepackage{algorithm, algpseudocode}
\usepackage{wrapfig}
\usepackage{listings}
\usepackage[separate-uncertainty=true,binary-units]{siunitx}
\usepackage{placeins}

\newcommand{\wpm}[1]{{\scriptstyle\color{darkgray} \pm #1}}  
\newcommand{\mbns}[1]{\textit{#1}}          
\newcommand{\mbf}[1]{\textbf{\mbns{#1}}}    

\newcommand{\dsmob}{MP:mob$\downarrow$}
\newcommand{\dsmp}{MP:sta$\uparrow$}
\newcommand{\dsgp}{GP:sta$\uparrow$}

\newcommand{\hide}[1]{}  
\newcommand{\note}[1]{}  
\newcommand{\sageev}[1]{}  

\newcommand{\Sv}{S\textsubscript{v}}
\newcommand{\apriori}{\textit{a priori}}

\newcommand{\etc}{\textit{etc.}}
\newcommand{\ie}{\textit{i.e.}}
\newcommand{\eg}{\textit{e.g.}}

\usepackage{xcolor}
\definecolor{halfgray}{gray}{0.55}
\definecolor{webgreen}{rgb}{0,.5,0}
\definecolor{webbrown}{rgb}{.6,0,0}
\definecolor{webblue}{HTML}{0000ED}

\definecolor{linkcolor}{HTML}{991408}  
\definecolor{citecolor}{HTML}{2E7E2A}  
\definecolor{filecolor}{HTML}{131877}  
\definecolor{menucolor}{HTML}{727500}  
\definecolor{runcolor} {HTML}{137776}  
\definecolor{urlcolor} {HTML}{0a2bbf}  

\hypersetup{%
    colorlinks=true,%
    ,linkcolor=linkcolor%
    ,citecolor=citecolor%
    ,filecolor=filecolor%
    ,menucolor=menucolor%
    ,runcolor=runcolor%
    ,urlcolor=urlcolor%
}

\def\keyFont{\fontsize{8}{11}\helveticabold }
\def\firstAuthorLast{Lowe {et~al.}} 
\def\Authors{
Scott~C.~Lowe\,$^{1,2,*}$,
Louise~P.~McGarry\,$^{3,*}$,
Jessica~Douglas\,$^{3}$,  
Jason~Newport\,$^{4,5}$, 
Sageev~Oore\,$^{1,2}$,
Christopher~Whidden\,$^{1,4}$,
and Daniel~J.~Hasselman\,$^{3}$
}
\def\Address{
$^{1}$Faculty of Computer Science, Dalhousie University, Halifax, Nova Scotia, Canada \\
$^{2}$Vector Institute, Toronto, Ontario, Canada \\
$^{3}$Fundy Ocean Research Centre for Energy, Dartmouth, Nova Scotia, Canada \\
$^{4}$DeepSense, Dalhousie University, Halifax, Nova Scotia, Canada \\
$^{5}$Marine Environmental Research Infrastructure for Data Integration and Application Network, Halifax, Nova Scotia, Canada
}
\def\corrAuthor{Scott C. Lowe}
\def\corrEmail{scott.lowe@dal.ca; scottclowe@gmail.com}

\begin{document}
\onecolumn
\firstpage{1}

\title[Echofilter]{Echofilter: A Deep Learning Segmentation Model Improves the Automation, Standardization, and Timeliness for Post-Processing Echosounder Data in Tidal Energy Streams}

\author[\firstAuthorLast ]{\Authors} 
\address{} 
\correspondance{} 

\extraAuth{Louise P. McGarry \\ louise.p.mcgarry@gmail.com}

\maketitle

\begin{abstract}

Understanding the abundance and distribution of fish in tidal energy streams is important for assessing the risks presented by the introduction of tidal energy devices into the habitat. However, tidal current flows suitable for tidal energy development are often highly turbulent and entrain air into the water, complicating the interpretation of echosounder data. The portion of the water column contaminated by returns from entrained air must be excluded from data used for biological analyses.
Application of a single algorithm to identify the depth-of-penetration of entrained air is insufficient for a boundary that is discontinuous, depth-dynamic, porous, and varies with tidal flow speed.

Using a case study at a tidal energy demonstration site in the Bay of Fundy, we describe the development and application of deep machine learning models with a U-Net based architecture that produce a pronounced and substantial improvement in the automated detection of the extent to which entrained air has penetrated the water column.

Our model, Echofilter, was found to be highly responsive to the dynamic range of turbulence conditions and sensitive to the fine-scale nuances in the boundary position, producing an entrained-air boundary line with an average error of 0.33\,m on mobile downfacing and 0.5--1.0\,m on stationary upfacing data, less than half that of existing algorithmic solutions.
The model's overall annotations had a high level of agreement with the human segmentation, with an intersection-over-union score of 99\% for mobile downfacing recordings and 92--95\% for stationary upfacing recordings.
This resulted in a 50\% reduction in the time required for manual edits when compared to the time required to manually edit the line placement produced by the currently available algorithms. Because of the improved initial automated placement, the implementation of the models permits an increase in the standardization and repeatability of line placement.

\tiny
 \keyFont{ \section{Keywords:}
    machine learning,
    deep learning,
    hydroacoustics,
    entrained air,
    marine renewable energy,
    tidal energy,
    environmental monitoring,
    marine technology
}
\end{abstract}

\section{Introduction}

The need for clean, non-carbon emitting, alternatives for power production is well established \citep{ipcc2021}. 
With advancements in technology, energy extraction from kinetic marine sources (ocean current, tidal energy streams, and wave) have recently emerged as potential contributions to the suite of renewable energies for the generation of electricity \citep{IRENA2020, Copping2020, Roberts2016, Cada2007}.
In the case of energy extraction from tidal energy streams, tidal turbines are introduced into nearshore, coastal ecosystems that can be important habitats of major biological importance to fish for migration, nursery, and feeding activities \citep{Tsitrin2022, Melvin2012, DFO2018, Blaber2000}.
The development of this nascent industry is therefore introducing new technologies and new but uncertain risks into the marine environment \citep{DFO2008}.
In regions where fish stocks are managed or listed for special protections, regulators require monitoring for potential effects for fish as a condition for licensing tidal and other marine renewable energy (MRE) projects.
Depending on local flow and bathymetric characteristics in the nearshore environments, the current flows suitable for tidal energy development can be turbulent \citep{Cornett2015,Melvin2015,Williamson2017,Perez2021,Wolf2022}, entraining persistent and deeply penetrating air into the water column.
An efficient backscatterer of sound, the presence of air in the water column complicates the post-processing activities for data collected with acoustic instruments.

Hydroacoustic methods, applied to data collected with scientifically calibrated echosounders, are used to quantify the distribution and abundance of fish in the marine environment \citep{Benoit-Bird, Fernandes2002, Johannesson1983}. 
Echosounders emit a pulse of sound (a ``ping'') into the water and record the magnitude of the returned backscatter (the ``echo'') \citep{Simmonds}. 
The advantage of echosounders is the ability to sample the full water column in high spatiotemporal resolution. 
However, to achieve the goals of biological analyses for fish presence and distribution, backscatter recorded from physical interfaces must be excluded, including from the seafloor or sea surface (sea-air interface) and those portions of the water contaminated by backscatter from entrained air.
The international standard solution to this is to use the software Echoview (Echoview Software Pty Ltd., Hobart, Australia), which enables advanced visualization and post-processing of hydroacoustic data.
Echoview includes a library of highly configurable, parameterized algorithms by which to achieve the work of post-processing, including defining the boundaries of the region suitable for biological analyses.

The classical algorithms of Echoview generally produce appropriate placement for the lines designating the seafloor and sea surface given their continuous, strongly reflective, and non-porous natures. 
In contrast, the boundary of the entrained-air penetration is indistinct, porous, and discontinuous, and formed of local features that can only be distinguished from biological features through their broader context.
The profile of entrained air is further complicated for recordings at sites where the penetration of entrained air is influenced by tidal flow speeds which can range from slack tide to \SI{5}{\metre\per\second} (\num{10}\,knots), \eg{} Bay of Fundy \citep{Karsten2013}.
These characteristics limit the potential for classical algorithms to successfully identify the extent of entrained air within the water column.
This lack of automation has important consequences for hydroacoustic data post-processing and analyses:
\begin{itemize}
    \item substantial and time-consuming manual edits are required to refine the ping-by-ping demarcation of the ambit of entrained air,
    \item the quantity of edits generates analyst fatigue putting at risk the regions where the full force of analyst attention is needed for discerning usable data, and
    \item standardization and/or repeatability is impossible to achieve between analysts and within the work of a single analyst.
\end{itemize}

Machine learning is a methodology that enables the construction of models through the use of example input/output data.
In particular, deep learning allows us to build the complex models which are necessary to solve challenging tasks which would otherwise require a human to laboriously perform \citep{deeplearning2015,Schmidhuber2015,Goodfellow-book}.
Deep learning models have revolutionized computer vision over the last 10 years \citep{alexnet,resnet,turinglecture2021}, have been successfully applied to image segmentation tasks \citep{unet,yolo,deep-segmentation-survey}, and have attained human-level or superhuman performance at narrow tasks \citep{KarpathyBlog2014,ImageNet2014,resnet,Santoro2016,AlphaZero}. 
We hypothesised that a deep neural network would be able to solve the task of placing the entrained-air line correctly.
Hence we deployed machine learning methods, with a convolutional neural network architecture inspired by U-Net \citep{unet} and EfficientNet \citep{efficientnet}, to determine whether such models can generate an entrained-air line with better placement than the existing classical algorithms (as implemented in Echoview) and thus reduce the amount of human labour needed to complete this task.

In the deep learning framework, an artificial neural network is instantiated with a particular architectural design (with randomly initialized parameters), and its parameters are iteratively updated through gradient descent in order to maximize performance at the objective task.
Through this training process, the network learns to approximate a function that maps a set of input stimuli to the correct outputs.
In the context of this work, the input to the model was a 2-D image-like representation of the hydroacoustic recording for which the axes are depth and time, and the intensity at each pixel is the volume backscattering strength (\Sv{} \si{\deci\bel} re: \SI{1}{\per\metre}); we refer to this input as an \textit{echogram}. 
The model's main output is a prediction of the depth of the entrained-air boundary line for each point in time (each ping).
In addition to this, our model also predicts the depths of the seafloor and sea surface boundary lines, and (for each datapoint) whether the echosounder was active (emitting pings) or passive (in listening-only mode).

Our final implementation, \textit{Echofilter}, is openly available under the \href{https://www.gnu.org/licenses/agpl-3.0.en.html}{AGPLv3} license.
Python source code and a stand-alone Windows executable are available at \url{https://github.com/DeepSenseCA/echofilter}, with command line interface (CLI) and application programming interface (API) documentation available at \url{https://DeepSenseCA.github.io/echofilter/}.

\section{Materials and Equipment}
\label{s:materials}

\subsection{Data sources}

Hydroacoustic data was collected from two tidal energy demonstration sites within the Bay of Fundy in Nova Scotia, Canada: Minas Passage in which flow speeds can exceed \SI{5}{\metre\per\second} \citep{Karsten2013} and Grand Passage in which flow speeds can achieve \SI{2.5}{\metre\per\second} \citep{Guerra}.

\begin{figure}[tb]
    \centering
    \includegraphics[width=\textwidth]{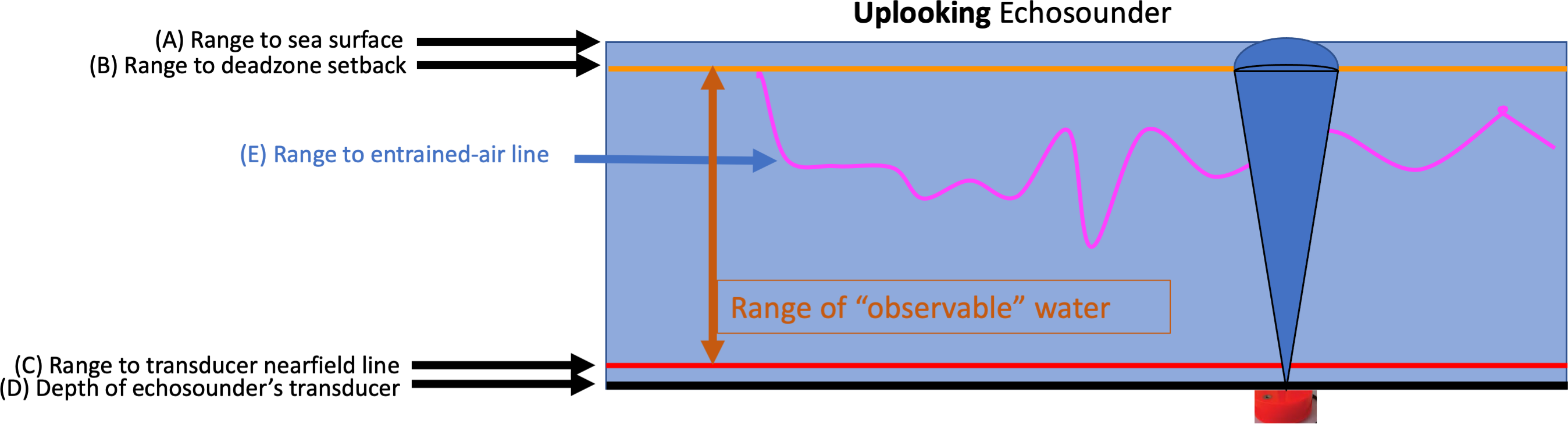}
    \caption{Illustration of the boundaries defining the range of the observable water column for an upfacing echosounder.
    \textbf{(A)}~Range to sea surface.
    \textbf{(B)}~Deadzone setback: the range in which biomass within the acoustic beam is indistinguishable from the strong surface echo \citep{Simmonds}.
    \textbf{(C)}~Nearfield line: a line set at a constant range to exclude the transducer ``nearfield'' in which the sound pulse is not yet organized \citep{Simmonds}.
    \textbf{(D)}~Position of the upfacing transducer.
    \textbf{(E)}~Entrained-air line: depth range of observable water is restricted to that not contaminated by entrained air.  }
    \label{fig:echosounder}
\end{figure}

``Stationary'' data was collected using a calibrated Simrad EK80 WBAT \SI{7}{\degree} split-beam echosounder operating in continuous wave (CW) mode at \SI{120}{\kilo\hertz} in Minas Passage and in Grand Passage.
The echosounder, with its transducer in an upward facing orientation was attached to a platform deployed to the seafloor (see \autoref{fig:echosounder}).
The seawater depth at the platform location varied with tide height from \SIrange{29}{44}{\meter} at Minas Passage, and \SIrange{14}{20}{\meter} at Grand Passage.
The echosounder was deployed in Minas Passage for three 2-month periods in 2018. Data was recorded for 5 minutes every half hour. Passive data collection with the echosounder in listening-only mode to document system self-noise and record levels of ambient sound present at \SI{120}{\kilo\hertz} was collected during two of the three deployments. There were two deployments of the echosounder in Grand Passage during late 2019 and early 2020. In both cases, the echosounder was deployed for less than 14 days. Data collection cycle in Grand Passage consisted of one-hour continuous data collection in alternating hours. Short durations of passive data were collected each hour.

``Mobile'' data was collected from the Minas Passage site using a calibrated Simrad EK80 WBT \SI{7}{\degree} split-beam echosounder operating in CW mode at \SI{120}{\kilo\hertz}. The transducer was deployed in a downward facing orientation attached via polemount to the vessel. The mobile survey pattern consisted of a set of six parallel transects, each of length \SI{1.8}{\kilo\metre} and separated by \SI{200}{\metre}, encompassing the Minas Passage multi-berth tidal energy demonstration site in the northern portion of the Passage, plus three reference transects located across the Passage near the southern shore. For ten of the seventeen mobile surveys, one additional transect was added to sample a region of interest in the demonstration site. The mobile surveys consisted of discrete 24-hour data collection periods during which the grid of transects was traversed four times, weather permitting. A completed grid consisted of one with-the-current and one against-the-current traverse of each transect. Seventeen such surveys were conducted between May 2016 and October 2018. Seawater depths ranged from \SIrange{13}{67}{\meter}. Mobile data collection included periods of passive data collection with the transiting of each transect. No mobile data was collected in Grand Passage.

The echosounder data files were imported into Echoview (version~10.0) and post-processed in the typical way: (i)~assigned calibration parameters, (ii)~examined the data and removed noise, (iii)~removed the passive data from further processing, (iv)~set a line at constant range from the transducer face (\SI{1.7}{\metre} in this case) by which to exclude the transducer nearfield and, (v)~applied Echoview algorithms to estimate, for each ping, the position of the seafloor (for downfacing echosounder) or sea surface (for upfacing echosounder) and the depth-of-penetration of the entrained air. In order to exclude the acoustic deadzone inherent in echosounder data \citep{Simmonds}, a one-meter offset was applied to the bounding line (seafloor or sea surface) and to the entrained-air line.

\subsection{Data partitioning}

\begin{table}[tb]
  \centering
  \caption{%
Summary of datasets used in this study: their recording locations, mobility, and recording orientation.
Additionally, we indicate the sizes of the dataset partitions, in terms of the number of contiguous recordings (duration dependent on dataset), and the total duration of the recordings measured in number of pings (k:~thousand; M:~million).
}
\label{tab:datasets}
\centerline{
\begin{tabular}{llllrrrrrr}
\toprule
        &               &           & & \multicolumn{3}{c}{\textnumero{} Recordings} & \multicolumn{3}{c}{\textnumero{} Pings} \\
\cmidrule(r){5-7} \cmidrule(l){8-10}
Dataset & Location      & Mobility  & Orientation              & Train & Val &Test& Train & Val & Test \\
\midrule
\dsmob{}& Minas Passage & Mobile    & Downfacing ($\downarrow$)&   727 &  91 & 245 & 1.21M & 148k & 394k \\
\dsmp{} & Minas Passage & Stationary& Upfacing ($\uparrow$)    & 7,249 & 919 & 875 & 2.45M & 305k & 300k \\
\dsgp{} & Grand Passage & Stationary& Upfacing ($\uparrow$)    &   118 &   0 &  28 & 0.36M &    0 &  96k \\
\bottomrule
\end{tabular}
}
\end{table}

The full suite of Echoview files were divided into sets of files for training, validating, and testing the machine-learning models.
The mobile downfacing dataset collected at Minas Passage consisted of 17 surveys, repeated at (a subset of) the same 10 transects on 17 different days spanning the course of three years.
We selected two transects and placed all recordings from these in the test set.
The remaining data was partitioned into training, validation, and (unused) ``test2'' partitions with an 80/10/10\% split, stratified against the season in which the data was collected (winter vs non-winter) to ensure an equal split of the sparser winter recordings.
The stationary data was grouped into blocks of 6h of consecutive recordings, and these blocks were partitioned at random (without stratification).
We placed 80\% of the \dsmp{} and \dsgp{} data files in the training partition.
For the \dsmp{} dataset, 10\% of the data was used for model validation and 10\% for final testing.
Due to its smaller size, we did not use any \dsgp{} data for the model validation process and kept the remaining 20\% of the data for testing.
The number of recordings and pings for each partition of each dataset is indicated in \autoref{tab:datasets}.

Files used for manual evaluation were selected from the \dsmp{} and \dsgp{} validation and test partitions, and chosen to ensure 24-hour coverage, with both neap tide and spring tide included.
Stratifying the samples in this way ensured we would see examples of model performance under best-case and worst-case entrained air scenarios.
We also selected files from the \dsmp{} and \dsgp{} training partitions to inspect, in order to determine whether errors were due to applying the models on new data, indicating an issue with the model's ability to generalize (overfitting), or whether there was also a problem on the training data, indicating an issue with the model design (underfitting).

\section{Methods}

\subsection{Echoview algorithm}

As a baseline to benchmark our models against, we used Echoview algorithms to generate seafloor (downfacing recordings), sea surface (upfacing recordings), and entrained-air boundary lines.

Two separate Echoview algorithmic approaches were used for estimating the ambit of entrained air within the stationary and mobile datasets, respectively.
For data collected during stationary surveys, periods of recorded passive-data were excluded and a 2-D Gaussian blur was applied to the remaining echogram using the Echoview ``XxY Convolution'' operator.
We used a 13-by-13 kernel, and a standard deviation of $\sigma = 2.0$ in both depth (over return sample indices) and time (over ping indices) dimensions.
\hide{A Gaussian blur, or Gaussian smoothing, is designed to reduce image noise and detail, and can be used to enhance image structure, using a normal distribution for calculating data values for the XxY sliding window.
The standard deviation, $\sigma$, specifies the strength of the blurring.
The kernel cell values were calculated using the equation for a Gaussian distribution in two dimensions,
\begin{equation}
\phi(u,v) = \frac{1}{2\pi \sigma^2}\exp\left(\frac{-(u^2+v^2)}{2\sigma^2}\right)
,\end{equation}
where $u$ is the kernel row index (relative to the center $[0,0]$) of the specified sliding window, $v$ is the kernel column index (relative to the center $[0,0]$) of the specified sliding window; and $\sigma$ is the standard deviation of the Gaussian distribution. 
The parameters used here were defined as follows: input samples per calculation was 169 (13 rows (samples) by 13 columns (pings)) and blurring was set to ``strong''; a value of $\sigma = 2.0$.
} 
We then used the ``Threshold Offset'' line picking operator, with a minimum threshold boundary of \SI{-80}{\deci\bel}, to search below the surface line and define the ambit of entrained air for each ping.
The position identified within each ping was then used as the automated demarcation between entrained air and water column in the original \Sv{} echogram.

For data collected during mobile surveys, the ``Best Bottom Candidate'' line picking operator was used to estimate the ambit of entrained air at each ping.
Because the Best Bottom Candidate operator identifies the first instance of strong signals deeper than weak signals in the water column, we first inverted the intensity of the \Sv{} echogram by multiplying the values by \num{-1} and adding \num{-150} to each result.
We then used the Best Bottom Candidate operator, parameterized with \SI{-70}{\deci\bel} for the minimum \Sv{} for a good pick and for the discrimination level, to identify the interface between entrained air (``weak'') values, and water column (``strong'') values in the inverse echogram.
The position identified for each ping was then used as the automated demarcation between entrained air and water column in the original \Sv{} echogram.
This standardized protocol was used for the last 8 mobile surveys (surveys 10 through 17).
For the first 9 mobile surveys, the protocol was inconsistent (sometimes including smoothing operations on the line, and offsets of varying sizes) and yielded variable outputs; these surveys were excluded from our benchmarking analysis described in \autoref{s:res-quantitative}.

The Best Bottom Candidate line picking operator was used to estimate the position of the seafloor (downfacing recordings), and the sea surface (upfacing recordings). For seafloor detection, the default settings were used and included a bottom offset of \SI{0.5}{\metre}, except for the first two of the seventeen mobile surveys for which some of the parameters were adjusted. For sea surface detection, the default settings were used except for the backstep discrimination level which was halved to \SI{-25}{\deci\bel}.

The entrained-air and seafloor lines produced by the Echoview algorithms were used as seed lines which expert human annotators, with reference to the \Sv{} echogram including a minimum \Sv{} threshold set to \SI{-66}{\deci\bel}, then manually adjusted to create corrected, finalized annotations.
These human-refined annotations were used as the targets for training the machine learning model.

\hide{
The Echoview-automated placement of the boundary of the entrained air as defined for both the stationary and mobile surveys was scrutinized and modified within the context of the \Sv{} echogram for which a minimum \Sv{} threshold of \SI{-66}{\deci\bel} was applied.
}

\subsection{Data preprocessing}
\label{s:preprocessing}

Annotated data was stored in Echoview EV files, which contain both the \Sv{} data and human-generated annotations for the boundary lines.
The EV files were opened in Python using \verb|win32com| to interface with Echoview's programming interface (API), and exported into several files.
The surface, seafloor, and entrained-air lines were exported into Echoview line (EVL) file format.
The \Sv{} data was exported into CSV format twice as follows.
The first \Sv{} CSV file (``raw \Sv{} CSV'') was exported with all EV exclusion settings disabled, and contained the entire \Sv{} data in the EV file.
The second \Sv{} CSV file (``clean \Sv{} CSV'') used the exclusion settings as implemented in the EV file so that all data which should be excluded from ecosystem analyses was masked out, appearing as the \verb|NaN| \href{https://support.echoview.com/WebHelp/Reference/File_formats/Export_file_formats/Special_Export_Values.htm}{indicator value} $-9.9\times10^{37}$ in the output CSV.
This means all datapoints above the entrained-air line, below the seafloor line (for downfacing recordings), passive data, bad data time periods where the analyst deemed a sequence of pings to contain data throughout the water column too contaminated by returns from entrained air or suspended sediment to use at all, and other miscellaneous localized ``bad data'' caused by anomalous events such as a rope drifting into view which the analyst had labelled for exclusion, were removed from the output (set to the \verb|NaN| indicator value), leaving only the datapoints deemed as ``good data'' by the analyst.

Since this export process requires using Echoview to read in and export the data from the EV file, and Echoview is only available for Windows, this first step of the data processing pipeline must be performed on a Windows system with a licensed copy of Echoview installed.
The remaining steps in the data processing and model training pipeline only require Python and can be run on any operating system.

The CSV files and EVL files were loaded into Python with a custom data loader.
The depth resolution (and number of datapoints) per ping sometimes differed during a recording session, resulting in data with an uneven sampling resolution; we addressed this by finding the modal depth resolution across pings and linearly interpolating the data for each ping onto the same array of depth sample points.
We created a ``target mask'' based on the location of \verb|NaN|-values in the clean \Sv{} CSV.
This target mask corresponds to the overall target for the network's output.
The depth lines loaded from the EVL files were linearly interpolated onto the same set of timestamps as the \Sv{} data.

We observed some discrepancies between the depth lines and the mask, which was caused by (1) off-by-one differences when the line threshold is applied in Echoview compared with our own interpolation of the line; (2) analysts using boxes or freehand regions to annotate exclusion regions which are adjacent to the boundary lines.
We handled this by identifying the upper and lower contiguous extent of the masked out area to generate new lines from the mask.
For the entrained-air line, we primarily used the deepest extent of the two options as provided via the line annotation and the mask annotation.
For the seafloor line, we primarily used the original line annotation as the network's output target, but we also produced a second line (with more aggressive removal) which extended higher up the water column to include any additional masked out area.
The spare ``aggressive'' version of the seafloor line was included as an auxiliary target during training.

The surface line annotations were mostly unchanged by the annotators from the output produced by Echoview's algorithms.
These were observed to be mostly accurate, but contained occasional large jumps in value.
These outliers were detected and removed by using a median filter as follows.
We applied a median filter with a kernel length of 201 and observed the residual between the raw signal and the median filter.
Values more than 5 standard deviations (robustly estimated from the interquartile range, $\sigma=\text{iqr}/1.35$) were set to the median value.
We then applied a median filter with a kernel length of 31 and removed anomalous values more than 4 standard deviations (robustly estimated from the interdecile range, $\sigma=\text{idr}/2.56$) from the median.
The second step was repeated until no anomalies were removed.
Additionally, if the surface line was ever deeper than the entrained-air line, we set it to be the same depth as the entrained-air line.
We found this anomaly removal process produced surface lines of sufficient quality.
For downfacing samples, the surface line was set at \SI{0}{\metre} (coincident with the transducer face).

Passive data annotations were taken as hard-coded on/off cycles where known \apriori.
Otherwise, passive data collection periods were identified using a bespoke algorithm.
The first \Sv{} responses, corresponding to depths closest to the echosounder, have large intensities when the echosounder is active and much lower values when the echosounder is passive.
We identified passive data periods by observing the first 38 depth sample points (after our interpolation step onto a common sampling grid).
We took the difference in \Sv{} between consecutive pings, and then the median across the first 38~depth samples for each ping.
Median differences which exceeded \SI{\pm25}{\deci\bel} were identified as boundary points between passive and active recording periods.

Bad data periods were identified as collections of consecutive pings for which all the data was masked out.
Periods of passive data recording were excluded from the bad data periods.
Bad data periods in which the entrained-air line was at or below the seafloor line throughout the entire period were also excluded.

Bad data patches were identified by the ``pixels'' in the echogram which were masked out for any reason not already covered by being above the entrained-air line, below the seafloor line, during a period of passive data collection, or during a period of time identified as a bad data period.

Our data was comprised of both upfacing and downfacing echosoundings.
In the recording data structure, and exported CSV files, the $y$-dimension is stored as increasing distance from the echosounder.
To standardize our inputs to the network, we flipped the orientation of the upfacing data such that increasing indices in the $y$-dimension corresponded to increasing depth within the water column.

The number of timepoints per file was much larger than we could reasonably supply to the network as a single input ``image''.
Moreover, it is important that a single training batch contains a diversity of training data.
To prevent the system from having to read in the contents of an entire recording file when needing to select only a small subset of the data to present for each training step, we broke the training data into chunks (shards) each with a length of 128 samples.

The pipeline for converting the CSV and EVL data into the preprocessed training shards can be executed with the command \verb|echofilter-generate-shards|.

\subsection{Training inputs}
\label{s:training-input}

When analysing echosounder data, it is common practice to offset the seafloor and entrained-air boundary lines by a fixed distance, \SI{1}{\metre} for the echosounders used here. The purpose of the fixed-distance offsets are to exclude those portions of the data near boundaries, such as the sea surface or seafloor or the entrained-air boundary, that may be biased due to the echosounder deadzone (see \autoref{fig:echosounder}) which is a function of the shape of the spherically spreading beam intersecting with a surface \citep{Simmonds}. In addition, it generates a buffer between the boundary of the entrained air and the data reserved for biological analyses, so as to exclude returns from entrained air adjacent to, but not connected to, the pronounced entrained-air boundary.
This ensures processing errs on the side of excluding slightly more data, instead of accidentally including bad data. 
Some datasets had an offset of \SI{1}{\metre} included in the line definitions, whereas others did not.
We standardized this by subtracting offsets from the lines which had them included.
Consequently, the model's target output is to predict the exact boundary locations, and offsets can be added to its outputs as appropriate via optional settings in the Echofilter API.

Each training input image was normalized independently, based on the distribution of \Sv{} values within the training input.
Normalization was performed by subtracting the median over all \Sv{} values, and dividing by a robust estimate of the standard deviation derived from the interdecile range ($\sigma= \text{idr}/2.56$).
A small number of \verb|NaN| values were present in the raw \Sv{} data, and these were set to a value of $-3$ after the normalization step.

The maximum apparent range of the echogram can in some cases be several times further than the actual depth of the water column.
This is because the depth dimension corresponds to the time-of-flight of the signals, the maximum of which is determined by a maximum range parameter chosen by the operator of the echosounder, which may be held the same across many recordings and thus may be much larger than the local depth of the water column.
In order to get the most precise output for the entrained-air lines from the trained model, we would like to zoom in on only the salient region of the image: the water column, extending from seafloor to sea surface.
This allows the model to predict the boundary point with sufficiently high granularity.
However, since the depth of the seafloor is not necessarily known \apriori{}, the model needs to be able to determine the depth of the seafloor, or range to sea surface, from the full echogram as well.
For testing, we thus use a two-step approach.
First, the full echogram is presented to the network and the seafloor and/or surface lines are predicted.
These outputs are used to zoom in on the water column.
Second, this zoomed-in echogram is presented to the network, and precise seafloor, surface, and entrained-air lines are generated.

Inputs to the network are samples from the distribution of plausible echograms.
During training, inputs to the network were drawn from the training partition and augmented with several operations.
(i)~Temporal stretching, stretch/squashed by a factor sampled log-uniformly from $[0.5, 2]$.
(ii)~Random depth cropping.
With $p=0.1$, the depth was left at the full, original extent.
With $p=0.1$, the echogram was zoomed in on the range from the shallowest surface depth to the deepest seafloor depth (the ``optimal'' zoom).
With $p=0.4$, the echogram was zoomed in to a random range of depths close to the optimal zoom, stretched or squashed by up to 25\%, but never so much as to remove more than 25\% of the entrained-air line or (for downfacing recordings) more than 50\% of the seafloor line.
With $p=0.4$, the echogram was zoomed in to a random range of depths between the full original extent and the ``optimal'' extent.
Depth upper and lower limits were selected uniformly across the appropriate range.
(iii)~Random reflection in the time (ping) dimension, performed with $p=0.5$.
(iv)~``Color'' jitter.
We applied a ``brightness'' augmentation by offsetting normalized \Sv{} values by a random additive offset chosen uniformly from $[-0.5,+0.5]$, and a ``contrast'' augmentation by multiplying normalised \Sv{} values by a random multiplicative factor chosen uniformly from $[0.7,1.3]$.
The same random offset and factor were used for each pixel in an echogram input.
The order of the brightness and contrast augmentations was randomly selected for each input.
(v)~Elastic grid deformation, performed with $p=0.5$.
Elastic deformation was performed separately in the depth and time dimensions, to create an elastic grid deformation.
We chose to deform the dimensions separately, instead of jointly as per a standard elastic deformation where space is stretched/squashed in a 2-D manner, because our targets are mostly at the ping level (depth of lines at each ping, whether the ping is passively or actively sampled, \etc{}) and apply to the entire column of data.
A standard 2-D elastic deformation would break the relationship between our input and target; performing a joint elastic deformation on the echogram input would make it challenging to relate the input to the targets.
We used $\sigma=8$ in the time dimension, $\sigma=16$ in the depth dimension, and $\alpha=0.1$ in both dimensions.
The echogram was interpolated in 2-D, with the interpolation order randomly selected from linear, quadratic, and cubic (equal weighting).

Finally, the echogram was rescaled to size $(128, 512)$ pixels (time-by-depth) for presentation to the network with nearest-neighbour interpolation. 


\subsection{Model architecture}

\begin{figure}[tb]
    \centerline{
    \includegraphics[width=0.98\textwidth]{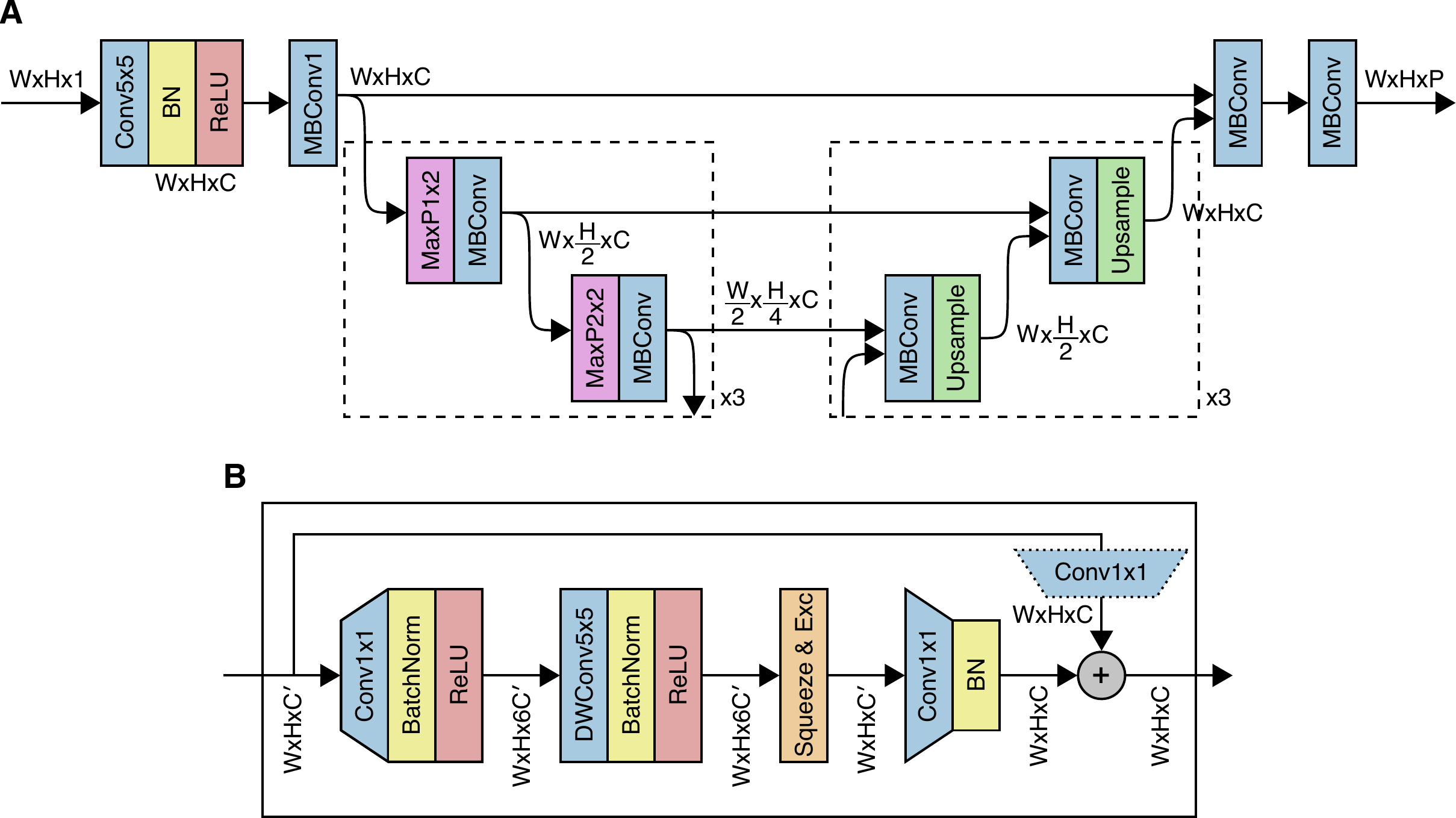}
    }
    \caption{Neural network architecture for the Echofilter model.
    \textbf{(A)} Main architecture, using an adaptation of the U-Net framework with 6 downsampling blocks, 6 upsampling units, and a single skip connection at each spatial resolution.
    The network contains a convolutional layer with $5\!\times\!5$ kernel and ``same'' padding (Conv5x5); BatchNorm (BN); rectified linear unit (ReLU); MBConv blocks (see panel B); max pooling (MaxP), with either $1\!\times\!2$ or $2\!\times\!2$ kernel and stride; and bilinear upscaling (Upsample) layers.
    The size of the latent representation of the image as it passes through the network is indicated. We train the network with $W\!=\!128$, $H\!=\!512$, $C\!=\!32$.
    \textbf{(B)} Structure of MBConv block, containing pointwise convolution (Conv1x1), depth-wise convolution (DWConv5x5) and Squeeze \& Excite layers.
    For downscaling blocks $C'\!=\!C$, and for upscaling blocks $C'\!=\!2C$.
    The pointwise convolution on the residual branch is only present for upscaling blocks, where $C'\!\neq\!C$.
    }
    \label{fig:arch}
    \label{fig:arch-main}
    \label{fig:mbconv}
\end{figure}

The model architecture used is a U-Net \citep{unet} with EfficientNet MBConv blocks \citep{mobilenet,efficientnet}, illustrated in \autoref{fig:arch}.
This architecture is a convolutional neural network (CNN) with residual skip connections across blocks, 6 encoder layers where the size is spatially compressed, 6 decoder layers where the size is expanded back to the original input dimensions, and skip connections from the encoder to decoder blocks.
The network has a backbone width of $32$ channels throughout, and each MBConv block is inverse residual with an expansion factor of 6 (except the very first block, with has an expansion factor of 1).
We used depthwise-separable convolutions with a kernel size of 5, and ReLU activations.
We used Squeeze \& Excite attention layers \citep{squeezeandexcitation} on each block with a reduction factor of 2.
In total, our final models each had 1.63M trainable parameters.

Since the input is rectangular, with higher resolution in the depth dimension, we downscaled the time dimension at a slower rate than the depth dimension.
Downscaling was performed with max-pooling using a kernel size and stride of either $1\times2$ or $2\times2$ (alternating blocks).
The depth dimension was downscaled after every block, whilst the time dimension was downscaled every other block.

The decoder branch was a mirror of the encoder: upscaling in the depth dimension after every block, and in the time dimension every other block.
Upscaling was performed using bilinear interpolation with \texttt{torch.nn.Upsample}.

The model has 10 output planes.
These correspond to the probability (represented in logit form) that a pixel is at the boundary point for:
(1)~the entrained air (expanded),
(2)~the entrained air (original),
(3)~the seafloor line (expanded),
(4)~the seafloor line (original),
(5)~the surface line;
and the probability (logit) that a pixel is within
(6)~a passive data period,
(7)~a bad data period (vertical region),
(8)~a miscellaneous bad data patch (to accompany expanded lines),
(9)~a miscellaneous bad data patch (to accompany original lines),
(10)~a miscellaneous bad data patch (to accompany original seafloor/expanded entrained air).

In practice, the expanded/original lines are almost identical and their pseudo-replication during training was superfluous, but their inclusion did indirectly increase the contribution of the entrained-air and seafloor lines towards the overall loss term.
When performing inference with the model, we discard outputs 2, 4, 9, and 10.

For the Bifacing model, these 10 output planes are replicated three times.
One is the standard output, the second are logits which are updated only on downfacing inputs, and the third are logits which are updated only on upfacing inputs.
In this way, the model learns to represent conditional probabilities $P(\text{boundary}|\text{upfacing})$, \etc{}
After training the model, we can ask it to predict the boundaries and masks agnostic of the orientation of the recording, or conditioned on the orientation (upfacing or downfacing).

\subsection{Model training}

The model was optimized with gradient descent to minimize a loss function.
The loss function acts as a proxy for the task of interest; a high loss corresponds to worse performance on the task, and a low loss to better performance.
We constructed our loss function as the sum of several terms, each corresponding to one of the output planes produced by the model.
The loss terms for the seafloor, sea surface, and entrained-air lines were each the cross-entropy between the column of logits across all depths for a single ping against a one-hot representation of the depth of the line.
The loss terms for the passive collection and bad data periods were binary cross-entropy between the model's output for that ping (a single scalar, after collapsing the depth-dimension with log-avg-exp; \citealp{logavgexp}) and the target value.
The loss term for the localized bad data regions was binary cross-entropy.
Outlying surface line values detected with our algorithm during preprocessing were masked out from the training objective.
We took the mean over pings for all loss terms.
We took the mean over the batch dimension; for outputs conditioned on the orientation of the echosounder, we masked out irrelevant samples before taking the batch-wise mean.
When training the bifacing model with conditioning signals, all stimulus presentations were double-counted and the entire loss was divided by two to correct for this.


The model was optimized using the RangerVA optimizer \citep{Ranger}, which combines RAdam, Lookahead, and gradient centralization \citep{lookahead,RAdam,GradientCentral,calibratingAGM}, with a weight decay of \num{1e-5}.
We used a batch size of 12 samples, and stratified the batches to contain the same ratio of downfacing and upfacing samples as available in the aggregated training set.
The learning rate (LR) followed a cyclic learning rate schedule \citep{clr,onecycle,superconvergence}.
In each cycle, the learning rate was warmed up for the first 10\% of training, held constant for 40\% of training, then warmed down for the last 50\%.
During the LR warmup period, the momentum was decreased from a maximum of $\beta_1=0.98$ to a base of $\beta_1=0.92$, and then increased back to 0.98 during the LR warmdown period.
Both the LR and momentum were increased and decreased with cosine annealing.
The second moment parameter was held constant at $\beta_2=0.999$ throughout training.
In the first cycle, the model was trained for 100 epochs with a maximum learning rate of $\text{LR}=0.012$.
In subsequent cycles, the training duration was progressively doubled and maximum learning rate halved.
We trained two models: the Bifacing model was trained for three cycles (700 epochs), whilst the Upfacing-only model was trained for two cycles (400 epochs).
The model parameters were saved at the end of each cycle for subsequent analysis.
We chose to stop the cyclic training process when the model's validation performance had reached a plateau.

The Upfacing model was trained on \dsmp{} and \dsgp{} datasets, which contain only upfacing \Sv{} recordings.
The Bifacing model was trained on the \dsmob{} dataset in addition to the \dsmp{} and \dsgp{} datasets.
To address the smaller size of the \dsgp{} dataset, we upsampled it by presenting echograms drawn from it twice per epoch instead of once (for both models).

The model architecture and training hyperparameters were each selected over a series of manual searches against the validation partition with short training durations of 5 or 20 epochs.

The network was trained using PyTorch~1.2.0 and CUDA~10.2.
The model training and testing were done on the DeepSense high performance computing cluster with each training cycle or test using a~20 Core IBM Power8NVL \SI{4.0}{\giga\hertz} compute node with \SI{512}{\giga\byte} of RAM and a pair of NVIDIA Tesla P100 GPUs with \SI{16}{\giga\byte} of GPU memory.

The Echofilter model can be trained using the command \verb|echofilter-train|, with training parameters set at the command prompt.

\subsection{Model output post-processing}
\label{s:meth:post}

The neural network model is configured to generate predictions for each output type at the pixel level.
That is to say, for each pixel in the input echogram, the network predicts a set of output variables at that particular pixel.
For the passive data and bad data periods, we convert this 2-D output into a 1-D time series by taking the log-avg-exp over the depth dimension \citep{logavgexp}.

We converted the model's output into lines as follows.
For each boundary line, our model predicts the probability that each pixel is the location of said boundary.
We integrated this probability across depth to create a cumulative probability density estimate, and identified the depth at which the cumulative probability exceeded 50\%.
In so doing, we generate a boundary depth prediction for every ping.

For the purposes of the machine learning model, all salient information needed to produce its outputs is contained in data at, or immediately surrounding, the water column.
However, some echosound recordings have much greater range than this, extending out beyond the water column with a large number of samples.
In order to put the echogram into the network, we scale the depth dimension down to 512~pixels.
For echograms much larger than the water column, this step incurs a loss of information, since the water column may occupy only a small fraction of the 512~pixel resolution.

In order to alleviate this issue, the Echofilter protocol may run the echogram through the network twice, once zoomed out and once zoomed in on the water column.
In the first instance, the echogram is ``zoomed out'' to the maximum extent and scaled down to 512~pixels.
The depth of the seafloor or sea surface line is noted (the choice of line depending on echosounder orientation), and used to estimate the extent of the water column.
Using a robust estimate of the standard deviation of depths in this line, we set our limit to be 4~standard deviations out from the mean of the line, or the furthest extent of the line, whichever is least distal.
For upfacing recordings, we zoom in on the range from the deepest recording up to this depth minus an additional \SI{2}{\metre}.
For downfacing recordings, we zoom in on the range from the shallowest recording depth down to this depth plus an additional \SI{2}{\metre}.
After cropping the echogram down to this range of depths, we scale it down to 512~pixels and present it to the network again.
The output from the second, ``zoomed-in'' presentation is used to determine the final entrained-air, surface/seafloor lines and other outputs.

This ``zoom+repeat'' technique provides gains (see \autoref{s:res-post}), but we expect it to be needlessly expensive when only a small fraction of the echogram is outside the water column.
For this reason, we only perform the second presentation if more than 35\% of the echogram would be cropped out.
This setting can be controlled with the \verb|--autozoom-threshold| argument to Echofilter.

In our analysis, we observed that Echofilter's predictions of the locations of ``bad data'' were not sufficiently accurate.
Furthermore, the mask can include a large number of small disconnected areas, which results in a inconveniently large number of regions to import into Echoview.
In order to counter this, we can merge together regions with small gaps in between them, and impose a minimum size threshold on regions to be included in the output.
We merged together consecutive passive regions annotations provided by the model with a gap smaller than 10 pings, and similarly for bad-data period labels.
Any remaining regions shorter than 10~pings in length were omitted from the final output.
For bad data patches, any patch with an area smaller than 25\,ping-metres was omitted from the final output.
In extremis, we can omit all bad-data annotations from Echofilter's region outputs.

An alternative solution to noisy outputs is to spatially smooth the output probabilities.
We can apply a Gaussian smoothing kernel across each output plane before converting the logits into probabilities, and subsequently into lines and regions.
However, we did not find this process yielded better results.

Lines and regions produced by Echofilter are exported into Echoview line (EVL) and region (EVR) files so they can be imported into Echoview.
Additionally, the Echofilter command line supports saving lines and regions directly into the EV file which it is processing (Windows~OS and a licensed copy of Echoview required), removing the subsequent step of manually importing the files.

Inference using a pretrained model can be performed on EV (Windows-only) or CSV files with the command \verb|echofilter|.
Pre-processing and post-processing options can configured be set at the command prompt.

\subsection{Performance quantification}

To compare the quality of the outputs from the Echoview algorithm and our Echofilter models, we used a selection of metrics to quantify their performances.
The results shown in \autoref{s:res-quantitative} were determined by using these performance measurements on the test partition of each dataset.
The test partition was neither seen by the model during training, nor used to optimize the model architecture and training process.

\subsubsection{Intersection-over-union}
\label{s:jaccard}
\label{s:IoU}

The model's output was evaluated using the intersection-over-union score (IoU), also known as the Jaccard index metric \citep{jaccard}, and Jaccard similarity coefficient score.
This metric is commonly used to evaluate the performance of image segmentation models within the field of computer vision.
The IoU of two masks is calculated by assessing their overlap; it is the ratio of the size of the intersection of the two masks against their union:
\begin{equation}
\operatorname{IoU}(\text{annotated}, \text{predicted}) = \frac{\operatorname{Area}(\text{annotated}\cap\text{predicted})}{\operatorname{Area}(\text{annotated}\cup\text{predicted})} \label{eq:jaccard}
.\end{equation}

\hide{
\begin{figure}[tb]
    \centering
    \includegraphics[width=.6\textwidth]{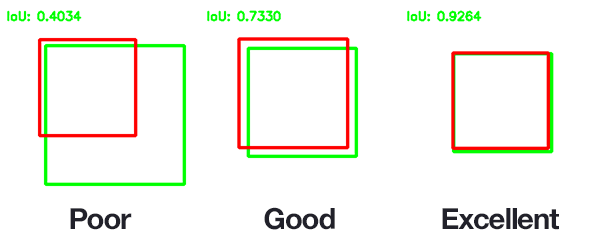}
    \caption{
    Example Jaccard index (IoU) calculation.
    \href{https://www.pyimagesearch.com/2016/11/07/intersection-over-union-iou-for-object-detection/}{``Intersection over Union''} by  Adrian Rosebrock, \href{https://creativecommons.org/licenses/by-sa/4.0/deed.en}{CC BY-SA 4.0}.
    }
    \label{fig:jaccard}
\end{figure}
}
For this study, one mask identifies the data marked as ``good'' by a human annotator, and the other mask is the data marked as ``good'' by the model.
A higher IoU is better, indicating the two masks are better aligned.
We chose to use this performance metric (instead of accuracy, \etc{}) because it is robust against padding the echogram with irrelevant range outside of the water column (below the seafloor for downfacing recordings, or above the sea surface for upfacing recordings).

For the \dsmp{} and \dsgp{} datasets, the IoU measurements we report are the total area of the mask intersections across the whole test set, divided by the total area of the union of the two masks (\ie{} the division operation performed after the summation).
For the \dsmob{} dataset, the IoU reported is the average IoU over all the EV files in the test set (\ie{} the division operation performed before the mean).
In both cases, we determine the standard error (SEM) by considering the distribution of IoU scores over EV files.
For any recording where the target mask is all marked as \verb|False| (no good data), the intersection of the predicted area with the target area is always 0, and any prediction from the model results in a anomalously minimal score.
Consequently, we excluded examples where the target was an empty mask when measuring the SEM.

\subsubsection{Mean Absolute Error}
\label{s:mae}

We performed further evaluation of the model's outputs using the mean absolute error (MAE) performance metric.
The MAE is defined as
\begin{equation}
\operatorname{MAE} = \frac{1}{n} \sum_{\forall i} | y_i - \hat{y}_i |
\label{eq:mae}
,\end{equation}
where $y_i$ is the target value for the $i$-th ping, $\hat{y}_i$ is the predicted value generated by the model, and $n$ is the number of pings to average over.
We applied the MAE to measure the quality of the output lines.
In this context, the MAE corresponds to the average distance (across pings) of the model's line from the target line.
A smaller MAE is better, indicating the model's line is (on average) closer to the target line.

When measuring the MAE of the lines, we excluded pings which were marked as being within a passive or bad data region in the target annotations.
To find the overall MAE, for \dsmob{} we determined the MAE within each file and averaged over files; for \dsmp{} and \dsgp{} we averaged over all pings across all files in the test set, weighting each ping equally.

Additionally, we report the standard error of the MAE.
This is determined by computing the MAE for each test file, and measuring the standard error across these independent measurements.

\subsubsection{Root-Mean-Square Error}
\label{s:rmse}

We measured the root-mean-square error (RMSE) in a similar manner to the MAE.
The RMSE is defined as
\begin{equation}
\operatorname{RMSE} = \sqrt{\frac{1}{n} \sum_{\forall i} (y_i - \hat{y}_i)^2 }
\label{eq:rmse}
,\end{equation}
where $y_i$ is the target value for the $i$-th ping, $\hat{y}_i$ is the predicted value generated by the model, and $n$ is the number of pings to average over.

When measuring the overall RMSE on the test data, we first found the average MSE over the whole test set, then took the square root.
Sample/file weighting, and standard error determination, was performed the same way as for the MAE.

\subsubsection{Cumulative error distribution}

We were particularly interested in how much human labour would be saved by the improvement in the annotations.
There clearly must exist some error threshold below which errors in the annotation have no significant impact on downstream analysis and hence do not need to be fixed by the analyst.
We speculate that this error tolerance threshold may be at around \SIrange{0.5}{1.0}{\metre}, since the lines are offset by \SI{1.0}{\metre} before performing biological analyses to ensure all ``bad data'' is excluded.
Hence we can crudely estimate what fraction of the model's output needs to be adjusted by the analyst by considering what fraction of pings are within \SI{0.5}{\metre} or \SI{1.0}{\metre} of the target line.

Since we can not be sure what the appropriate error tolerance is --- and the tolerable error threshold may vary depending on the application --- we can evaluate the performance of the model over a range of potential tolerance values by considering the cumulative distribution of the absolute error.
Such a plot shows the fraction of outputs which are within a certain absolute error threshold, and is similar to a receiver operating characteristic (ROC) curve.
If we seek to optimize this curve without assigning any particular error tolerance threshold, we can consider the total area above the curve (the expected rejection rate over all error tolerance thresholds), which we seek to minimise.
This area is precisely equal to the MAE metric.

\subsubsection{Test data weighting}
\label{s:test-method}

For the \dsmob{} dataset, test recordings were taken from two held out transects --- no recordings from these transects were presented during training.
This allows us to evaluate the performance of the model at novel recording locations which the model has not seen before.
Unfortunately, the protocol for annotating the seafloor was not consistent for the first 9 of the 17 \dsmob{} surveys; hence we evaluated the seafloor line and overall IoU only on test data from the final 8 surveys.
The entrained air, passive data collection, bad data periods, and bad data patches were evaluated on all 17 of the \dsmob{} surveys.

For the \dsmp{} and \dsgp{} datasets, our target surface lines were generated with the Echoview surface line detector, followed by automated anomaly detection, as described in \autoref{s:preprocessing}.
However, in some cases the Echoview algorithm fatally failed to detect the water--surface boundary, placing the surface line impossibly close to the echosounder, or impossibly far away.
When evaluating the surface line on the test set, we dropped recordings where the ``target'' surface line depth was outside the known range of low to high tide water depths for that recording site (Minas Passage: \SIrange{28.5}{44}{\meter}; Grand Passage: \SIrange{13.5}{20}{\meter}).
This allowed us to evaluate the Echofilter model's surface line predictions against sane target values, but this selectively removed almost all the occasions where the Echoview algorithm's predictions were wrong, severely compromising our ability to evaluate the performance of the Echoview algorithm at generating the surface line.

\section{Results}
\label{s:results}

We measured the performance of our models using coarse-grained quantitative metrics (\autoref{s:res-quantitative}), and compared to the output of algorithms built into Echoview as a baseline.
To further contextualise the level of performance attained by our models, we measured the level of agreement between expert annotators separately annotating the data (\autoref{s:annotator-agreement}).
We also evaluated the performance by detailed investigation with qualitative outputs (\autoref{s:res-qualitative}), and finally we evaluated the practical output of the model by measuring the amount of time taken to audit and correct the model output (\autoref{s:timesaving}).

\subsection{Quantitative evaluation}
\label{s:res-quantitative}

We evaluated the overall performance of our final model by comparing the final ``good data'' mask produced by the model with that of the target labels.
The target mask indicates which values within the echogram should be included in biological analyses.
This mask excludes all values above the entrained-air line, below the seafloor line, during passive data collection regions, or marked as ``bad data''.
Our model produces outputs corresponding to each of these elements, and combining these outputs allows us to generate a final output mask.
We measured the alignment between the two masks using the Intersection-over-Union (IoU), described in \autoref{s:jaccard}.

For our purposes, the most important output from the Echofilter model was the entrained-air line, which provides segmentation between the air entrained into the water column, and the rest of the water column.
To provide a human-interpretable measurement of the error in the placement of this line, we measured the mean absolute error (MAE) and root-mean-square error (RMSE) between the depth of model's entrained-air line and the target.
See \autoref{s:mae} and \autoref{s:rmse} for more details.

Other outputs from the model were evaluated similarly, using the IoU, MAE, and/or RMSE.
In all cases, we show the performance of the model on the test partition, which was held out during all stages of model development and training.

\subsubsection{Performance break-down across outputs}

We investigated the performance of the final Upfacing (@400 epoch) and Bifacing (@700 epoch) models across all outputs produced by the network, and compared the quality of these outputs against the Echoview algorithm.
The results were evaluated against the target annotations produced by a human expert, except for the surface line where the target was taken from the line produced by the Echoview algorithm but with anomalous values rejected (see \autoref{s:test-method}).

\begin{table}[tbh]
  \centering
  \caption{%
Final model performance (agreement with manual annotation) for each output.
The performance of the final Upfacing (@400 epoch) and Bifacing (@700 epoch) models, with thresholded zoom+repeat, merging/ignoring small output regions; compared against the performance of the Echoview algorithm as a baseline.
Bold: best model.
Italic: no significant difference from best (two-sided Wilcoxon signed-rank test, $p\!>\!0.05$).
}
\label{tab:test-perf-per-output}
\centerline{
  \scalebox{0.80}{
\begin{tabular}{lrrrrrrrr}
\toprule
                        & \multicolumn{2}{c}{\dsmob{}} & \multicolumn{3}{c}{\dsmp{}} & \multicolumn{3}{c}{\dsgp{}} \\
\cmidrule(l){2-3} \cmidrule(l){4-6} \cmidrule(l){7-9}
                        &                         & \multicolumn{1}{c}{Echofilter}&                 & \multicolumn{1}{c}{Echofilter} & \multicolumn{1}{c}{Echofilter} &     & \multicolumn{1}{c}{Echofilter} & \multicolumn{1}{c}{Echofilter} \\
Output                  & \multicolumn{1}{c}{Echoview} & \multicolumn{1}{c}{Bifacing} & \multicolumn{1}{c}{Echoview} & \multicolumn{1}{c}{Upfacing} & \multicolumn{1}{c}{Bifacing} & \multicolumn{1}{c}{Echoview} & \multicolumn{1}{c}{Upfacing} & \multicolumn{1}{c}{Bifacing} \\
\midrule
                        & \multicolumn{8}{c}{Intersection-over-Union (\%; larger is better)} \\
\cmidrule(l){2-3} \cmidrule(l){4-6} \cmidrule(l){7-9}
Overall                 &$     { 96.80}\wpm{0.34}$&$ \mbf{ 99.15}\wpm{0.08}$&$     { 90.41}\wpm{0.76}$&$ \mbf{ 95.08}\wpm{0.34}$&$\mbns{ 94.91}\wpm{0.35}$&$     { 87.66}\wpm{1.05}$&$\mbns{ 92.10}\wpm{1.00}$&$ \mbf{ 92.97}\wpm{1.00}$\\
Entrained-air           &$     { 97.37}\wpm{0.31}$&$ \mbf{ 99.11}\wpm{0.09}$&$     { 91.63}\wpm{0.72}$&$ \mbf{ 96.05}\wpm{0.29}$&$     { 95.96}\wpm{0.28}$&$     { 89.06}\wpm{1.03}$&$\mbns{ 94.49}\wpm{0.50}$&$ \mbf{ 94.95}\wpm{0.29}$\\
Surface                 & --                      & --                      &$ \mbf{ 99.83}\wpm{0.05}$&$\mbns{ 99.82}\wpm{0.02}$&$ \mbf{ 99.83}\wpm{0.02}$&--\hide{$ \mbf{100.00}\wpm{0.00}$}&$\mbns{ 98.59}\wpm{1.22}$&$ \mbf{ 99.86}\wpm{0.01}$\\
Seafloor                &$     { 99.33}\wpm{0.08}$&$ \mbf{ 99.79}\wpm{0.03}$& --                      & --                      & --                      & --                      & --                      & --                      \\
Air--Seafloor           &$     { 96.81}\wpm{0.34}$&$ \mbf{ 99.16}\wpm{0.08}$& --                      & --                      & --                      & --                      & --                      & --                      \\
Passive                 & --                      &$     { 99.78}\wpm{0.06}$& --                      &$ \mbf{100.00}\wpm{0.00}$&$ \mbf{100.00}\wpm{0.00}$& --                      &$\mbns{ 99.97}\wpm{0.01}$&$ \mbf{100.00}\wpm{0.00}$\\
Bad data period         & --                      & --                      & --                      &$ \mbf{ 40.58}\wpm{7.64}$&$\mbns{ 38.92}\wpm{7.42}$& --                      &$\mbns{ 24.68}\wpm{7.65}$&$ \mbf{ 25.78}\wpm{8.07}$\\
Patch (anomaly)         & --                      &$  0.00\wpm{0.00}$       & --                      &$ \mbf{  0.30}\wpm{0.12}$&$ \mbf{  0.30}\wpm{0.11}$& --                      &$ \mbf{  0.20}\wpm{0.07}$&$ \mbf{  0.20}\wpm{0.07}$\\
\midrule
                        & \multicolumn{8}{c}{Mean Absolute Error (\si{m}; smaller is better)} \\
\cmidrule(l){2-3} \cmidrule(l){4-6} \cmidrule(l){7-9}
Entrained-air           &$     {1.178}\wpm{0.295}$&$ \mbf{0.325}\wpm{0.031}$&$     {2.187}\wpm{0.147}$&$ \mbf{0.981}\wpm{0.044}$&$     {1.005}\wpm{0.045}$&$     {1.252}\wpm{0.198}$&$\mbns{0.577}\wpm{0.074}$&$ \mbf{0.532}\wpm{0.031}$\\
Surface                 & --                      & --                      &$ \mbf{0.062}\wpm{0.018}$&$\mbns{0.063}\wpm{0.007}$&$\mbf{0.062}\wpm{0.007}$&--\hide{$ \mbf{0.001}\wpm{0.000}$}&$\mbns{0.235}\wpm{0.232}$&$ \mbf{0.024}\wpm{0.002}$\\
Seafloor                &$     {0.279}\wpm{0.032}$&$ \mbf{0.089}\wpm{0.012}$& --                      & --                      & --                      & --                      & --                      & --                      \\
\midrule
                        & \multicolumn{8}{c}{Root-Mean-Square Error (\si{m}; smaller is better)} \\
\cmidrule(l){2-3} \cmidrule(l){4-6} \cmidrule(l){7-9}
Entrained-air           &$     {6.436}\wpm{0.390}$&$ \mbf{1.281}\wpm{0.064}$&$     {4.275}\wpm{0.196}$&$ \mbf{2.181}\wpm{0.085}$&$     {2.228}\wpm{0.088}$&$     {2.995}\wpm{0.508}$&$\mbns{1.244}\wpm{0.137}$&$ \mbf{1.104}\wpm{0.060}$\\
Surface                 & --                      & --                      &$     {1.323}\wpm{0.140}$&$     {0.149}\wpm{0.020}$&$ \mbf{0.134}\wpm{0.019}$&--\hide{$\mbf{0.019}\wpm{0.004}$}&$\mbns{3.019}\wpm{1.073}$&$ \mbf{0.035}\wpm{0.003}$\\
Seafloor                &$     {2.017}\wpm{0.187}$&$ \mbf{0.292}\wpm{0.024}$& --                      & --                      & --                      & --                      & --                      & --                      \\
\midrule
                        & \multicolumn{8}{c}{Proportion of pings where line placed within \SI{0.5}{m} of target (\%; larger is better)} \\
\cmidrule(l){2-3} \cmidrule(l){4-6} \cmidrule(l){7-9}
Entrained-air           &$     { 71.24}\wpm{1.60}$&$ \mbf{ 88.30}\wpm{0.82}$&$     { 52.67}\wpm{2.09}$&$ \mbf{ 61.28}\wpm{1.26}$&$\mbns{ 61.11}\wpm{1.27}$&$     { 61.18}\wpm{2.58}$&$\mbns{ 69.85}\wpm{2.54}$&$ \mbf{ 70.17}\wpm{1.47}$\\
Surface                 & --                      & --                      &$     { 99.39}\wpm{0.15}$&$\mbns{ 99.67}\wpm{0.27}$&$ \mbf{ 99.68}\wpm{0.26}$&--\hide{$\mbns{ 99.96}\wpm{0.01}$}&$\mbns{ 99.44}\wpm{0.57}$&$ \mbf{ 99.98}\wpm{0.01}$\\
Seafloor                &$     { 89.27}\wpm{1.02}$&$ \mbf{ 97.26}\wpm{0.70}$& --                      & --                      & --                      & --                      & --                      & --                      \\
\midrule
                        & \multicolumn{8}{c}{Proportion of pings where line placed within \SI{1.0}{m} of target (\%; larger is better)} \\
\cmidrule(l){2-3} \cmidrule(l){4-6} \cmidrule(l){7-9}
Entrained-air           &$     { 80.20}\wpm{1.33}$&$ \mbf{ 93.09}\wpm{0.59}$&$     { 58.34}\wpm{1.99}$&$ \mbf{ 74.76}\wpm{0.97}$&$     { 74.33}\wpm{1.01}$&$     { 69.10}\wpm{2.70}$&$\mbns{ 83.70}\wpm{1.84}$&$ \mbf{ 84.93}\wpm{0.92}$\\
Surface                 & --                      & --                      &$     { 99.44}\wpm{0.15}$&$\mbns{ 99.87}\wpm{0.15}$&$ \mbf{ 99.88}\wpm{0.15}$&--\hide{$ \mbf{100.00}\wpm{0.00}$}&$\mbns{ 99.46}\wpm{0.57}$&$ \mbf{100.00}\wpm{0.00}$\\
Seafloor                &$     { 95.30}\wpm{0.49}$&$ \mbf{ 98.75}\wpm{0.53}$& --                      & --                      & --                      & --                      & --                      & --                      \\
\midrule
                        & \multicolumn{8}{c}{Proportion of pings where line placed within \SI{2.0}{m} of target (\%; larger is better)} \\
\cmidrule(l){2-3} \cmidrule(l){4-6} \cmidrule(l){7-9}
Entrained-air           &$     { 87.58}\wpm{1.01}$&$ \mbf{ 96.46}\wpm{0.38}$&$     { 68.08}\wpm{1.80}$&$ \mbf{ 86.50}\wpm{0.66}$&$     { 86.12}\wpm{0.69}$&$     { 80.61}\wpm{2.26}$&$\mbns{ 93.33}\wpm{1.17}$&$ \mbf{ 94.20}\wpm{0.45}$\\
Surface                 & --                      & --                      &$     { 99.52}\wpm{0.15}$&$\mbns{ 99.93}\wpm{0.11}$&$ \mbf{ 99.94}\wpm{0.10}$&--\hide{$ \mbf{100.00}\wpm{0.00}$}&$\mbns{ 99.46}\wpm{0.57}$&$ \mbf{100.00}\wpm{0.00}$\\
Seafloor                &$     { 98.48}\wpm{0.18}$&$ \mbf{ 99.78}\wpm{0.16}$& --                      & --                      & --                      & --                      & --                      & --                      \\
\bottomrule
\end{tabular}
  }
}
\end{table}

We considered the IoU for each output, the results for which are shown in \autoref{tab:test-perf-per-output}.
The overall IoU compares the overall mask produced by removing pixels above the entrained-air line, below the seafloor line (if downfacing), during periods marked as passive data collection, and bad data annotations; we compare the mask obtained with the model against that from human annotation.
For the entrained-air line, the IoU measurement considers the area beneath the entrained-air line --- for upfacing recordings this extends to the echosounder, and for downfacing recordings it extends to the seafloor line provided by the expert's annotation.
Similarly the IoU measurement for the surface line extends from the surface to the echosounder, and is only measured for upfacing recordings.
For the seafloor line, we compare the area from the seafloor line to the echosounder.
For passive data region annotations, we compare the set of pings identified as passive by the model with the target annotations, performing a 1-D IoU calculation.
The vertical bad data periods are measured in the same way as the passive data region annotations, using a 1-D IoU.
The IoU for the bad data patches is a comparison of the area marked as bad data by the model with a target mask indicating the locations of bad data patches.

We found the entrained-air and seafloor boundaries produced by both models had statistically significantly higher agreement with the human annotation than the lines produced by the Echoview algorithm (two-sided Wilcoxon signed-rank test, $p\!<\!0.05$).
There was no significant difference between the outputs from the two models, except from the entrained-air line on \dsmp{} where the magnitude of the difference in performance was small.
Correspondingly, the quality of the overall output of the models were not significantly different from each other, but were significantly better than the Echoview algorithm.

The passive region annotations are highly accurate, reaching 100\% accuracy on \dsmp{} and \dsgp{}.
On the \dsmob{} dataset, the Bifacing model attains an IoU of 99.78\%.

The bad data period annotations were challenging for the model to replicate, attaining an IoU of only 40\% on \dsmp{} and 25\% on \dsgp{}.
The anomalous bad data patches were impossible for the network to learn with any meaningful reliability, with an IoU of $\leq\!0.3\%$.
The poor performance of both of these annotations yields an increase in performance when small outputs are ignored (as seen in \autoref{s:res-post}).
On \dsgp{}, the bad data period annotations are sufficiently poor to yield an increase in performance when they are dropped entirely (see \autoref{s:res-post}).

We measured the mean absolute error (MAE) of the entrained-air, surface, and seafloor lines (described in \autoref{s:mae}).
As shown in \autoref{tab:test-perf-per-output}, we found that the Bifacing model placed the entrained-air line on \dsmob{} with only \SI{0.325}{\metre} average error --- a significant reduction (two-sided Wilcoxon signed-rank test, $p\!<\!0.05$) over the Echoview algorithm, which had over three times as much error on average.
For the \dsmp{} and \dsgp{} datasets, the two Echofilter models had comparable performance, a significant reduction in error against the Echoview algorithm baseline which had more than twice the error of Echofilter on both the upfacing, stationary datasets.
Our findings when evaluating the entrained-air lines using the RMSE metric, and the proportion of pings within fixed distances of the target lines, were the same as with MAE.

The larger absolute error on the \dsgp{} dataset (\SIrange{0.53}{0.58}{\meter}) and \dsmp{} dataset (\SI{1.0}{\meter}) is indicative of the increased difficulty intrinsic to these datasets collected at sites where the persistence, depth-of-penetration, variability of depth-of-penetration, and proportion of water column contaminated by entrained air exceeded that typically found at the transect locations sampled by the \dsmob{} surveys.
In particular, we note that the average and standard deviation of the depth-of-penetration for the entrained air was \SI{12.4\pm 4.8}{\meter} for \dsmp{}, \SI{6.7 \pm 2.1}{\meter} for \dsgp{}, but only \SI{2.4 \pm 2.2}{\meter} for \dsmob{}.
This corresponds to \SI{34 \pm 13}{\percent} of the water column for \dsmp{}, \SI{41 \pm 12}{\percent} for \dsgp{}, but only \SI{6.2 \pm 5.3}{\percent} for \dsmob{}.

\begin{figure}[tb]
\centering
\includegraphics[width=.98\textwidth]{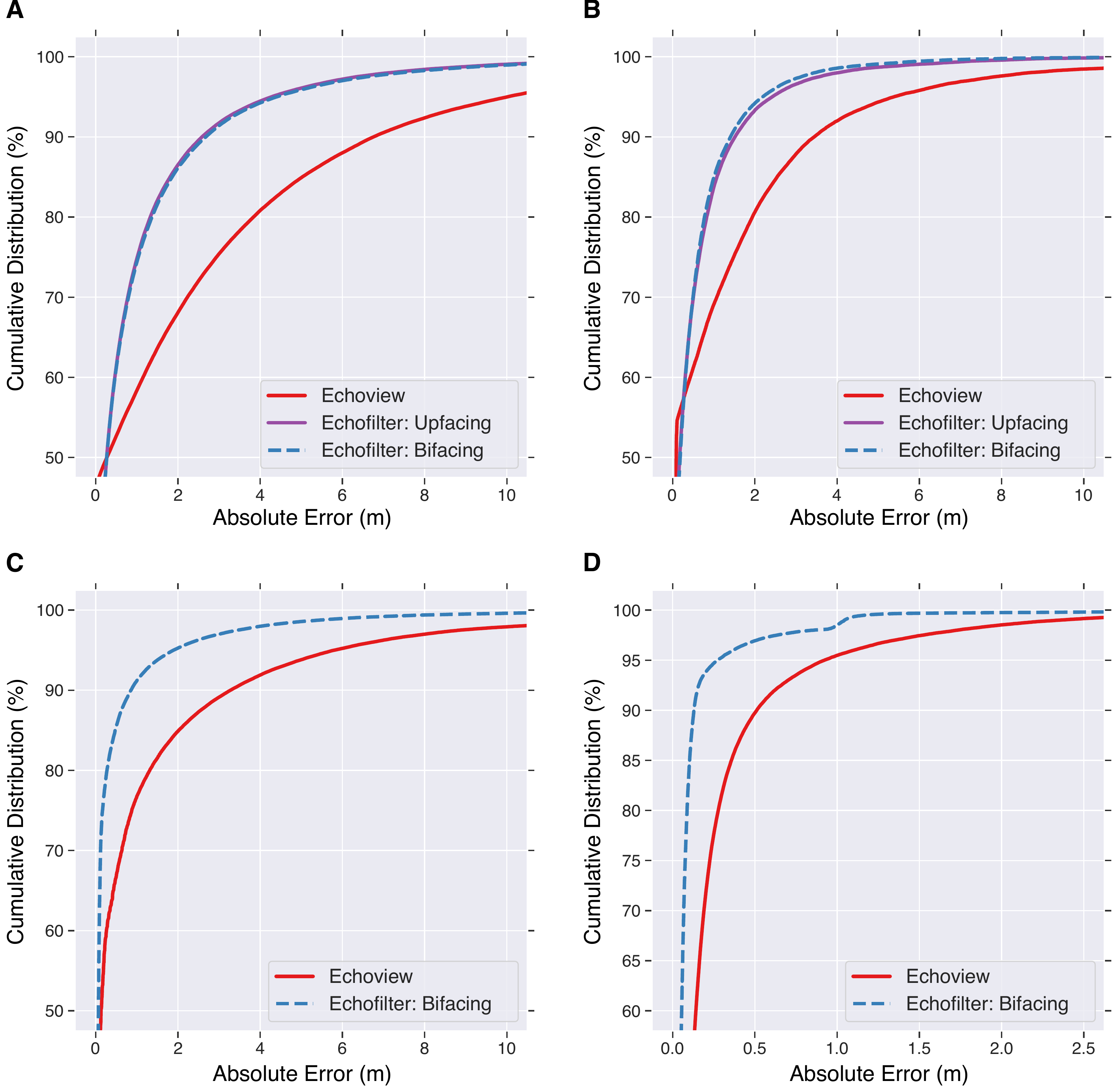}
\caption{
Cumulative distribution for the absolute error of entrained-air (\textbf{A}--\textbf{C}) and seafloor (\textbf{D}) lines generated by the models (Echoview: red; Upfacing@400ep: magenta; Bifacing@700ep: blue).
This indicates (on the y-axis) the fraction of pings where the output line was within a given threshold distance (x-axis) of the target line.
(\textbf{A}) Error in entrained-air line on \dsmp{}.
(\textbf{B}) Error in entrained-air line on \dsgp{}.
(\textbf{C}) Error in entrained-air line on \dsmob{}.
(\textbf{D}) Error in seafloor line on \dsmob{}.
}
\label{fig:cumulative-dist}
\end{figure}

As shown in \autoref{fig:cumulative-dist}A--C, the cumulative error distribution curve produced by Echoview is dominated by Echofilter for all values in the range of interest.
Echoview appears to outperform Echofilter when using very narrow error thresholds (error \SI{<0.2}{\metre}; the easiest 50\% of the data), however this is an artifact of the manual data annotation process, in which Echoview was used to generate initial annotations which were then corrected as needed by a human expert.

As shown in \autoref{tab:test-perf-per-output}, we found that the Bifacing model placed the seafloor line with very low error (only \SI{0.09}{\metre} on average) on the \dsmob{} test set.
Again, this was significantly lower than the Echoview algorithm, for which the MAE was three times higher (\SI{0.28}{\metre}).
Similar results were seen with the RMSE and proportion of pings within a tolerance threshold.
As shown in \autoref{fig:cumulative-dist}, the Echofilter cumulative error distribution for the seafloor line dominates Echoview.
We note there is a step jump in Echofilter model performance at \SI{1.0}{\metre} error; this is caused by inconsistencies in the training data annotation in the first 9 surveys which impacted the model fit (evaluated only on the last 8 surveys).

For the surface line annotation, we find that the Echofilter models have a MAE comparable to Echoview on \dsmp{} and outperform Echoview when considering the RMSE and fraction of pings within \SIrange{0.5}{2.0}{\metre} error threshold.
This is because the Echofilter models do not produce the anomalous surface line depths seen with the Echoview line generation, which we removed from our training and target lines.
On \dsgp{}, the Bifacing model produced better surface lines than the Upfacing model.
Manual inspection of the results demonstrates that the Upfacing model is sometimes confused by reflections in the additional range of these recordings (following the erroneous training targets generated by Echoview), whilst the Bifacing model was not confused by these reflections.

\subsubsection{Impact of post-processing steps}
\label{s:res-post}

We evaluated the performance of the final Upfacing and Bifacing models before and after each post-processing step impact described in \autoref{s:meth:post}. 
Our results are shown in \autoref{tab:test-perf-postprocessing}.

\begin{table}[tbh]
  \centering
  \caption{%
Impact of post-processing steps on the model performance metrics.
Bold: best pre-processing option.
Italic: no significant difference from best (two-sided Wilcoxon signed-rank test, $p\!>\!0.05$).
}
\label{tab:test-perf-postprocessing}
\centerline{
  \scalebox{0.975}{
\begin{tabular}{lrrrrrr}
\toprule
                        & \multicolumn{3}{c}{Overall IoU (\%)} & \multicolumn{3}{c}{Entrained-air MAE (\si{m})} \\
\cmidrule(r){2-4} \cmidrule(l){5-7}
Model                   & \multicolumn{1}{c}{\dsmob{}} & \multicolumn{1}{c}{\dsmp{}} & \multicolumn{1}{c}{\dsgp{}} & \multicolumn{1}{c}{\dsmob{}} & \multicolumn{1}{c}{\dsmp{}} & \multicolumn{1}{c}{\dsgp{}} \\
\midrule
Echoview algorithm      & $ 96.80\wpm{0.34}$      &$ 90.41\wpm{0.76}$       &$ 87.66\wpm{1.05}$       & $1.178\wpm{0.295}$      &$2.187\wpm{0.147}$       &$1.252\wpm{0.198}$       \\
\midrule
Upfacing w/o zoom       & --                      &$     { 95.06}\wpm{0.34}$&$     { 88.06}\wpm{3.88}$& --                      &$     {0.987}\wpm{0.045}$&$     {0.629}\wpm{0.076}$\\  
w/ zoom+repeat          & --                      &$ \mbf{ 95.11}\wpm{0.35}$&$\mbns{ 92.09}\wpm{1.01}$& --                      &$ \mbf{0.950}\wpm{0.041}$&$ \mbf{0.574}\wpm{0.071}$\\  
w/ thresholded z+r      & --                      &$     { 94.27}\wpm{0.46}$&$\mbns{ 92.07}\wpm{1.01}$& --                      &$     {0.981}\wpm{0.044}$&$\mbns{0.577}\wpm{0.074}$\\  
+ ignore small regions  & --                      &$     { 95.08}\wpm{0.34}$&$\mbns{ 92.10}\wpm{1.00}$& --                      &$     {0.981}\wpm{0.044}$&$\mbns{0.577}\wpm{0.074}$\\  
+ ignore all ``bad data''& --                     &$     { 94.77}\wpm{0.44}$&$ \mbf{ 93.01}\wpm{0.76}$& --                      &$     {0.981}\wpm{0.044}$&$\mbns{0.577}\wpm{0.074}$\\  
+ logit smoothing       & --                      &$     { 94.27}\wpm{0.46}$&$\mbns{ 92.48}\wpm{0.86}$& --                      &$     {1.099}\wpm{0.046}$&$\mbns{0.623}\wpm{0.095}$\\  
\midrule
Bifacing w/o zoom       &$     { 98.59}\wpm{0.09}$&$ \mbf{ 94.90}\wpm{0.35}$&$     { 88.35}\wpm{3.93}$&$     {0.402}\wpm{0.030}$&$     {1.004}\wpm{0.045}$&$     {0.589}\wpm{0.047}$\\  
w/ zoom+repeat          &$ \mbf{ 99.16}\wpm{0.08}$&$     { 94.86}\wpm{0.40}$&$\mbns{ 92.95}\wpm{1.01}$&$ \mbf{0.325}\wpm{0.031}$&$ \mbf{0.979}\wpm{0.044}$&$ \mbf{0.532}\wpm{0.031}$\\  
w/ thresholded z+r      &$ \mbf{ 99.16}\wpm{0.08}$&$ \mbf{ 94.90}\wpm{0.35}$&$\mbns{ 92.95}\wpm{1.01}$&$ \mbf{0.325}\wpm{0.031}$&$     {1.005}\wpm{0.045}$&$ \mbf{0.532}\wpm{0.031}$\\  
+ ignore small regions  &$\mbns{ 99.15}\wpm{0.08}$&$ \mbf{ 94.91}\wpm{0.35}$&$\mbns{ 92.97}\wpm{1.00}$&$ \mbf{0.325}\wpm{0.031}$&$     {1.005}\wpm{0.045}$&$ \mbf{0.532}\wpm{0.031}$\\  
+ ignore all ``bad data''&$\mbns{ 99.15}\wpm{0.08}$&$\mbns{ 94.74}\wpm{0.42}$&$ \mbf{ 93.45}\wpm{0.64}$&$ \mbf{0.325}\wpm{0.031}$&$     {1.005}\wpm{0.045}$&$ \mbf{0.532}\wpm{0.031}$\\  
+ logit smoothing       &$     { 98.90}\wpm{0.08}$&$     { 94.35}\wpm{0.43}$&$\mbns{ 93.12}\wpm{0.67}$&$     {0.385}\wpm{0.030}$&$     {1.103}\wpm{0.051}$&$\mbns{0.570}\wpm{0.047}$\\  
\bottomrule
\end{tabular}
  }
}
\end{table}

Compared with applying the model only once on the full echogram, using the two step ``zoom+repeat'' stimulus presentation provided a statistically significant increase in the entrained-air line placement as evaluated by the MAE (two-sided Wilcoxon signed-rank test, $p\!<\!0.05$) on all datasets, and for both the Upfacing and Bifacing model.
The overall IoU also significantly increased, except for the Bifacing model on \dsmp{} where ``zoom+repeat'' caused a very small, but statistically significant, decrease.

We also considered using a threshold of 0.35 to determine when to do the zoom+repeat step, following our assumption that a second application of the model on a zoomed-in echogram is not necessary when less than 35\% of the echogram data is outside the surface--seafloor extent.
We found that using this threshold had no impact on the performance of the models on the \dsmob{} and \dsgp{} datasets, where the range of the data extended far outside the surface--seafloor extent and hence zoom+repeat was \textit{de facto} always applied.
On the \dsmp{} dataset, where the recording range was not much further than the distance from seafloor to sea surface, there was a significant decrease in performance when a threshold was used to determine when to apply a second round of the model.
This suggests that the zoom+repeat protocol should always be used in order to yield the best annotation with the model.
Nonetheless, the rest of our results present in this paper use the (faster to perform) thresholded zoom+repeat, with a threshold of 0.35.

The remaining optional post-processing steps were considered with thresholded zoom+repeat in place.
We found no significant differences in the overall IoU when small regions were merged together or dropped from the output (changing the way regions are handled has no effect on the entrained-air line placement).
Omitting bad data regions and patches entirely had a positive impact on the overall performance on the \dsgp{} data, but a negative impact on \dsmp{} data.
This was because the bad data period predictions (as seen in \autoref{tab:test-perf-per-output}) were notably worse on \dsgp{} than \dsmp{}.
There was no impact on \dsmob{} data because the model did not predict any bad data regions on this test dataset.

We considered the effect of logit smoothing on the model's final output by applying this postprocessing step, in addition to thresholded zoom+repeat and ignoring all bad data annotations, with a Gaussian kernel size of 1.
We found that logit smoothing had a significant negative impact on the accuracy of the entrained-air line placement, and on the overall mask output, for all datasets.

\subsubsection{Impact of model training duration}

We investigated the impact of training time on the final model outputs.
We compared the output of each of the models at the end of each stage of the cyclic training process.
For this analysis, we used thresholded zoom+repeat, and merged/ignored small regions in the model output.

As shown in \autoref{tab:test-perf-training-cycles}, we found that further training cycles improved the performance on \dsmp{} and \dsmob{}, though with diminishing returns.
Additional training  \textit{reduced} the performance on \dsgp{}, but the reduction was not statistically significant.

\begin{table}[tb]
  \centering
  \caption{%
Performance of models after each training cycle (different total training durations).
Bold: best training duration.
Italic: no significant difference from best (two-sided Wilcoxon signed-rank test, $p\!>\!0.05$).
}
\label{tab:test-perf-training-cycles}
\centerline{
  \scalebox{0.975}{
\begin{tabular}{llrrrrrr}
\toprule
         &         & \multicolumn{3}{c}{Overall IoU (\%)} & \multicolumn{3}{c}{Entrained-air MAE (\si{m})} \\
\cmidrule(r){3-5} \cmidrule(l){6-8}
Model    &         & \multicolumn{1}{c}{\dsmob{}} & \multicolumn{1}{c}{\dsmp{}} & \multicolumn{1}{c}{\dsgp{}} & \multicolumn{1}{c}{\dsmob{}} & \multicolumn{1}{c}{\dsmp{}} & \multicolumn{1}{c}{\dsgp{}} \\
\midrule
Upfacing & 100ep   & --                      &$\mbns{ 95.05}\wpm{0.33}$&$ \mbf{ 93.32}\wpm{0.87}$& --                      &$     {1.001}\wpm{0.046}$&$ \mbf{0.520}\wpm{0.033}$ \\
         & 400ep   & --                      &$ \mbf{ 95.08}\wpm{0.34}$&$\mbns{ 92.10}\wpm{1.00}$& --                      &$ \mbf{0.981}\wpm{0.044}$&$\mbns{0.577}\wpm{0.074}$ \\
\midrule
Bifacing & 100ep   &$     { 98.93}\wpm{0.10}$&$     { 94.93}\wpm{0.32}$&$ \mbf{ 93.52}\wpm{0.69}$&$     {0.369}\wpm{0.034}$&$     {1.036}\wpm{0.047}$&$ \mbf{0.513}\wpm{0.032}$ \\
         & 400ep   &$     { 99.02}\wpm{0.09}$&$ \mbf{ 94.97}\wpm{0.33}$&$\mbns{ 93.18}\wpm{0.92}$&$     {0.329}\wpm{0.028}$&$     {1.022}\wpm{0.047}$&$\mbns{0.520}\wpm{0.032}$ \\
         & 700ep   &$ \mbf{ 99.15}\wpm{0.08}$&$     { 94.91}\wpm{0.35}$&$\mbns{ 92.97}\wpm{1.00}$&$ \mbf{0.325}\wpm{0.031}$&$ \mbf{1.005}\wpm{0.045}$&$\mbns{0.532}\wpm{0.031}$ \\
\bottomrule
\end{tabular}
  }
}
\end{table}

\subsection{Inter-annotator agreement benchmarking}
\label{s:annotator-agreement}

The extent to which air is entrained in the water column is not observed directly, and can only be estimated based on the echosounder recordings.
With training and experience, human annotators can learn which datapoints correspond to entrained air and which to fish populations within the water column.
However, without a ground truth measurement, the annotations are subjective and will differ between annotators.

With this in mind, it is difficult to know how well we could expect an ideal model to perform at the task.
It is infeasible to expect perfect agreement between the model and the human annotations, since human annotators do not always agree amongst each other and are not necessarily consistent in their choice of line placement.
We endeavoured to quantify how well our model performs by measuring the agreement between two human annotators, which acts as a baseline to estimate the Bayes error rate.

\begin{table}[tb]
  \centering
  \caption{%
Comparison of agreement between several annotation sources.
We compared several annotation methods against expert labels created by JD.
The intersection-over-union (IoU) across all recordings is shown, in addition to the mean absolute error (MAE) and root-mean-square error (RMSE) for the placement of the entrained-air separation line ($n=10$, $\pm$ inter-recording standard error).
Note that JD used Echoview to generate seed annotations for refinement into finalized annotations.
Bold: model with best agreement with target (JD) annotations.
Italic: no significant difference from best (two-sided Wilcoxon signed-rank test, $p\!>\!0.05$).
}
\label{tab:test-xcompare}
\centerline{
\begin{tabular}{lrrrrrr}
\toprule
                           & \multicolumn{4}{c}{IoU (\%; larger is better)} & \multicolumn{2}{c}{$\Delta$ Entrained-air (\si{m})} \\
\cmidrule(r){2-5}
\cmidrule(l){6-7}
Annotator                  & \multicolumn{1}{c}{Overall} & \multicolumn{1}{c}{Entrained-air} & \multicolumn{1}{c}{Bad data period} & \multicolumn{1}{c}{Patch} & \multicolumn{1}{c}{MAE} & \multicolumn{1}{c}{RMSE} \\
\midrule
Human expert (LPM)         &$\mbns{ 90.7}\wpm{1.2}$&$\mbns{ 92.4}\wpm{0.9}$&$ \mbf{ 97.8}\wpm{25.8}$&$     {  0.29}\wpm{0.10}$&$\mbns{0.86}\wpm{0.10}$&$\mbns{1.63}\wpm{0.17}$\\
\midrule
Echoview                   &$\mbns{ 88.7}\wpm{1.4}$&$     { 90.9}\wpm{1.1}$& --                     & --                      &$     {1.05}\wpm{0.12}$&$     {2.01}\wpm{0.19}$\\
Echofilter: Upfacing       &$\mbns{ 90.5}\wpm{2.9}$&$ \mbf{ 93.2}\wpm{0.9}$&$\mbns{ 70.7}\wpm{24.0}$&$\mbns{  0.32}\wpm{0.09}$&$ \mbf{0.76}\wpm{0.05}$&$ \mbf{1.25}\wpm{0.11}$\\
Echofilter: Bifacing       &$ \mbf{ 91.3}\wpm{1.3}$&$\mbns{ 93.0}\wpm{1.0}$&$\mbns{ 92.8}\wpm{25.5}$&$ \mbf{  0.38}\wpm{0.12}$&$\mbns{0.78}\wpm{0.04}$&$\mbns{1.27}\wpm{0.09}$\\
\bottomrule
\end{tabular}
}
\end{table}

We selected 10 EV~files from the Grand Passage stationary-upfacing dataset (\dsgp{}), ensuring that the selected files were composed of sufficiently complex data so that any differences in line placement between each annotator would be highlighted.
Annotations were generated by Echoview using the preexisting workflow\hide{ (\autoref{fig:dataflow}, left panel)}.
The Echoview annotations were edited independently by JD and LPM in order to create two sets of finalized annotations for all 10 files.
We then created annotations using Echofilter (models Upfacing@400ep and Bifacing@700ep, using thresholded zoom+repeat, and dropping small regions).
While both annotators are experts in this field, JD was the most experienced at handling this data --- her annotations constituted the majority of the annotations used to train the models.
Consequently, we treated JD's annotations as the ground truth labels, and measured the performance of the other annotation methods in comparison to her labels.

As shown in \autoref{tab:test-xcompare}, we found that the level of agreement in placement of the entrained-air line between Echofilter and JD exceeded that of LPM, with higher IoU and a smaller average distance between the line depths, though the difference was not statistically significant ($p\!>\!0.05$).
This suggests our model outputs have an accuracy comparable to human-level performance at this task.

\subsection{Manual evaluation of model outputs}
\label{s:res-qualitative}

Manual investigation of the Echofilter results were carried out by JD and LPM on a Windows 10 operating system, using Echoview 10, or Echoview 11 newly released at the time of testing.
The performance of Echofilter was evaluated on 24 Echoview files, selected from the test partition as described in \autoref{s:materials}.

\begin{figure}[!tbp]
    \centering
    \includegraphics[width=\textwidth]{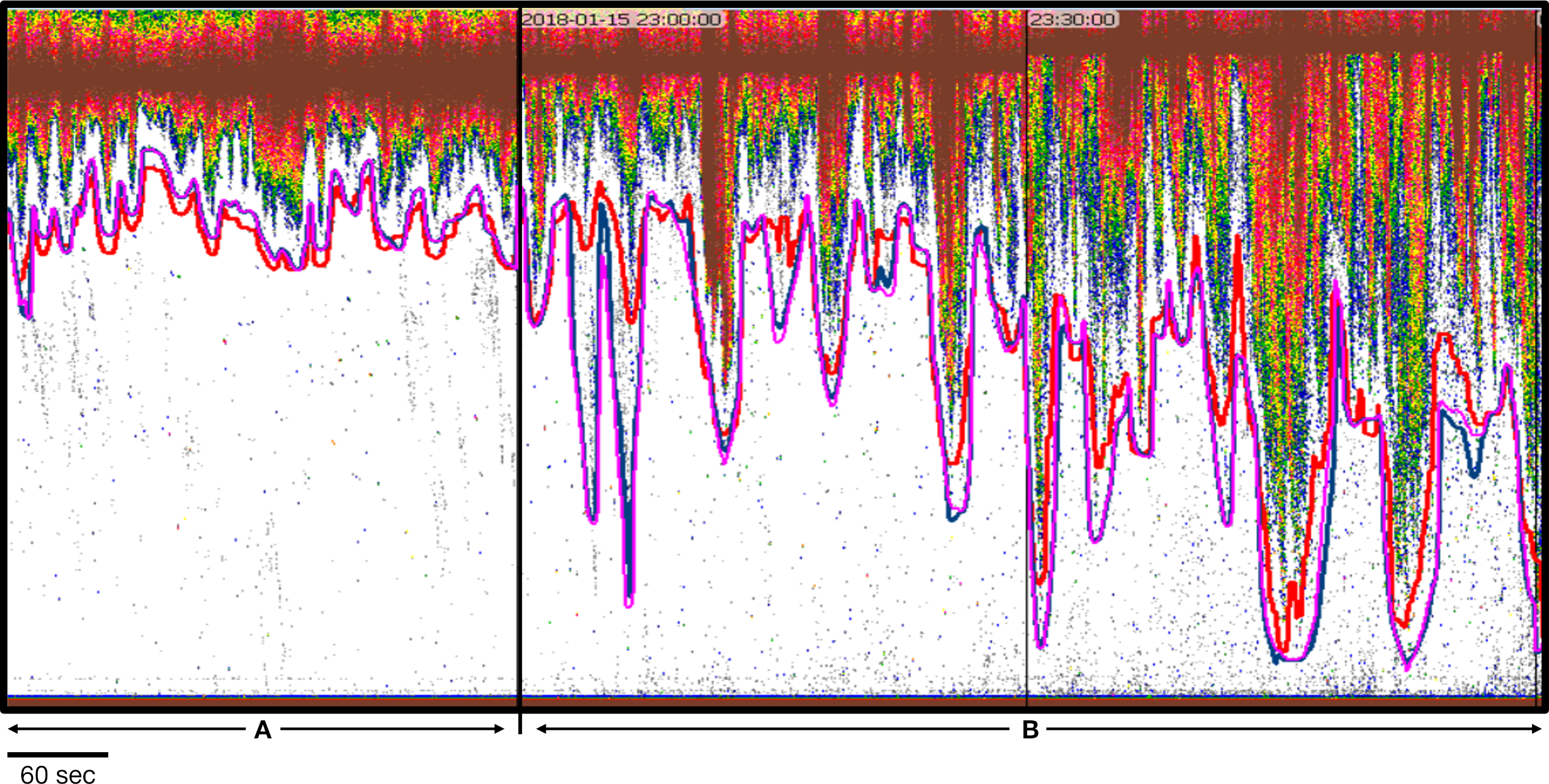}
    \caption{Entrained-air lines as defined by Echofilter (Upfacing@100ep: pink; Bifacing@100ep: blue) and by Echoview (red). \textbf{(A)} A 5-minute data collection period during which entrained air does not penetrate deeply into the water. The Echoview line is further from the entrained air than the Echofilter lines, leaving greater amounts of white space and thereby unnecessarily excluding more water column from analyses. \textbf{(B)} Two 5-minute data collection periods during which the returns from entrained air are more depth dynamic. The Echofilter placement of the entrained-air lines more closely reflect the penetration of the entrained air in terms of depth and width. In the horizontal dimension, the Echofilter lines are appropriately placed further from the entrained air in the particularly steep sections.  Note that Echofilter entrained-air lines as defined by each model (Bifacing@100ep and Upfacing@100ep) are essentially equivalent although not identical. }
    \label{fig:turbulenceline}
\end{figure}

\begin{figure}[!tbp]
    \centering
    \includegraphics[width=\textwidth]{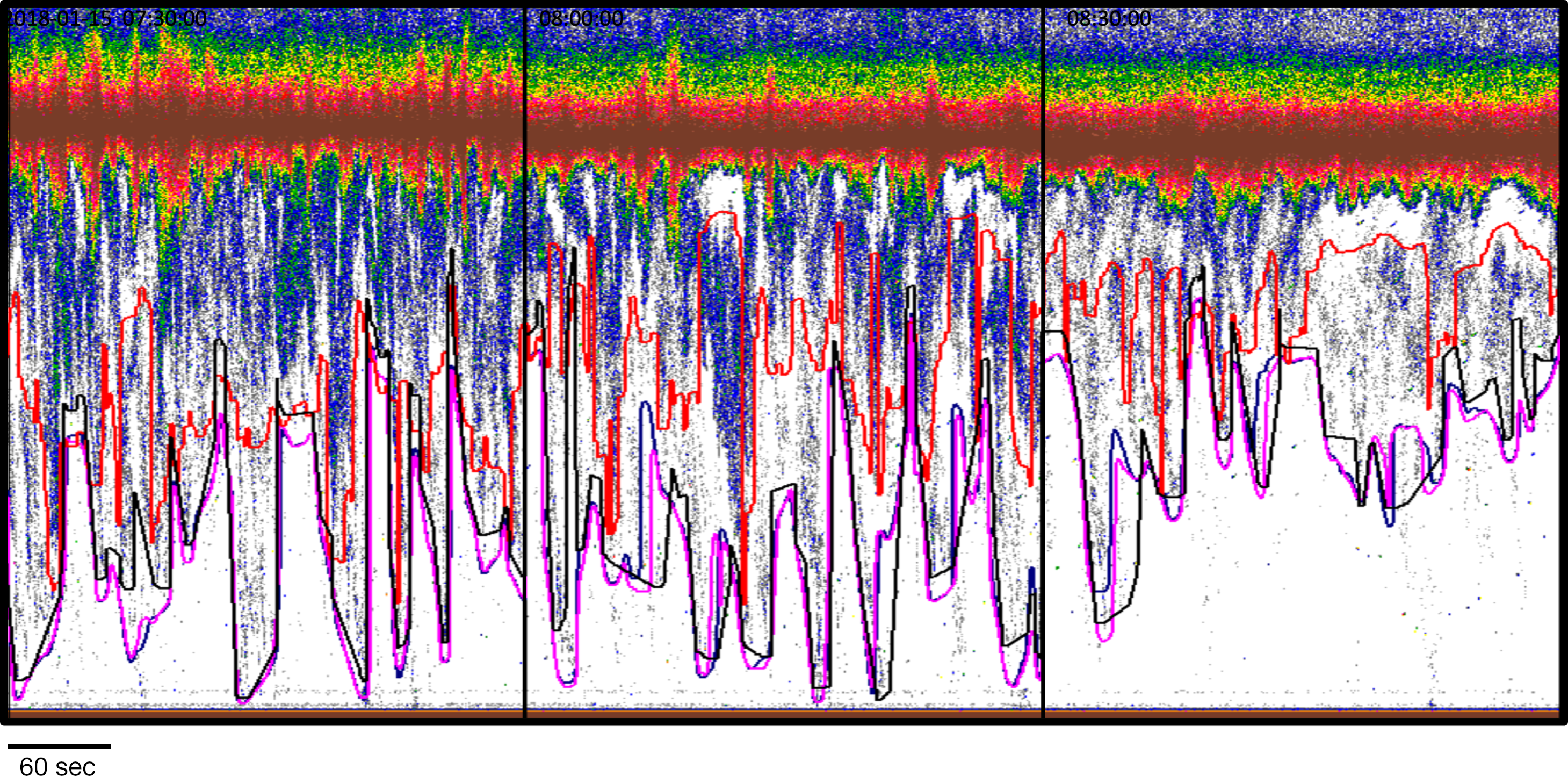}
    \caption{Echogram demonstrating that the entrained-air line as calculated by the two Echofilter models (Upfacing@400ep: pink; Bifacing@700ep: blue) is a pronounced improvement over that produced by Echoview (red line), and much closer to the target line created by the analyst (black).
    Data: stationary data with echosounder in upfacing orientation, recorded for 5 minutes every half hour at the Minas Passage site.}
    \label{fig:echofilterwins}
\end{figure}

\begin{figure}[!tbp]
    \centering
    \includegraphics[width=\textwidth]{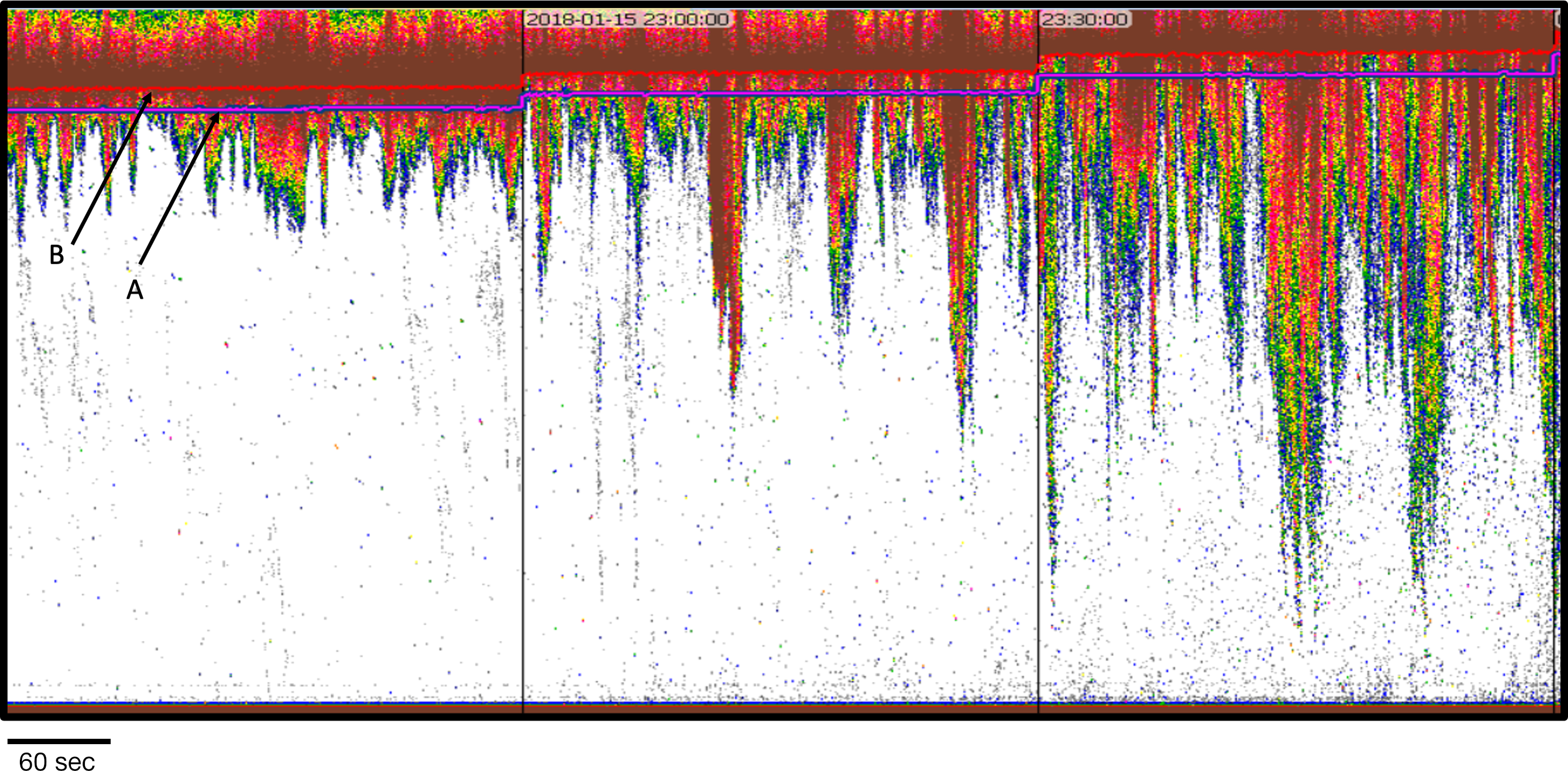}
    \caption{Example showing appropriate and adequate placement of the surface line by Echofilter models: Upfacing@100ep (pink) and Bifacing@100ep (blue) indicated at \textbf{A}. The line placements include the \SI{1}{\metre} offset required to eliminate bias from acoustic beam deadzone. For reference, the surface line, without offset, as defined by Echoview is also shown (red; line \textbf{B}).
    Data: stationary data with echosounder in upfacing orientation, recorded for 5 minutes every half hour at the Minas Passage site.}
    \label{fig:surfaceline}
\end{figure}

\begin{figure}[!tbp]
    \centering
    \includegraphics[width=\textwidth]{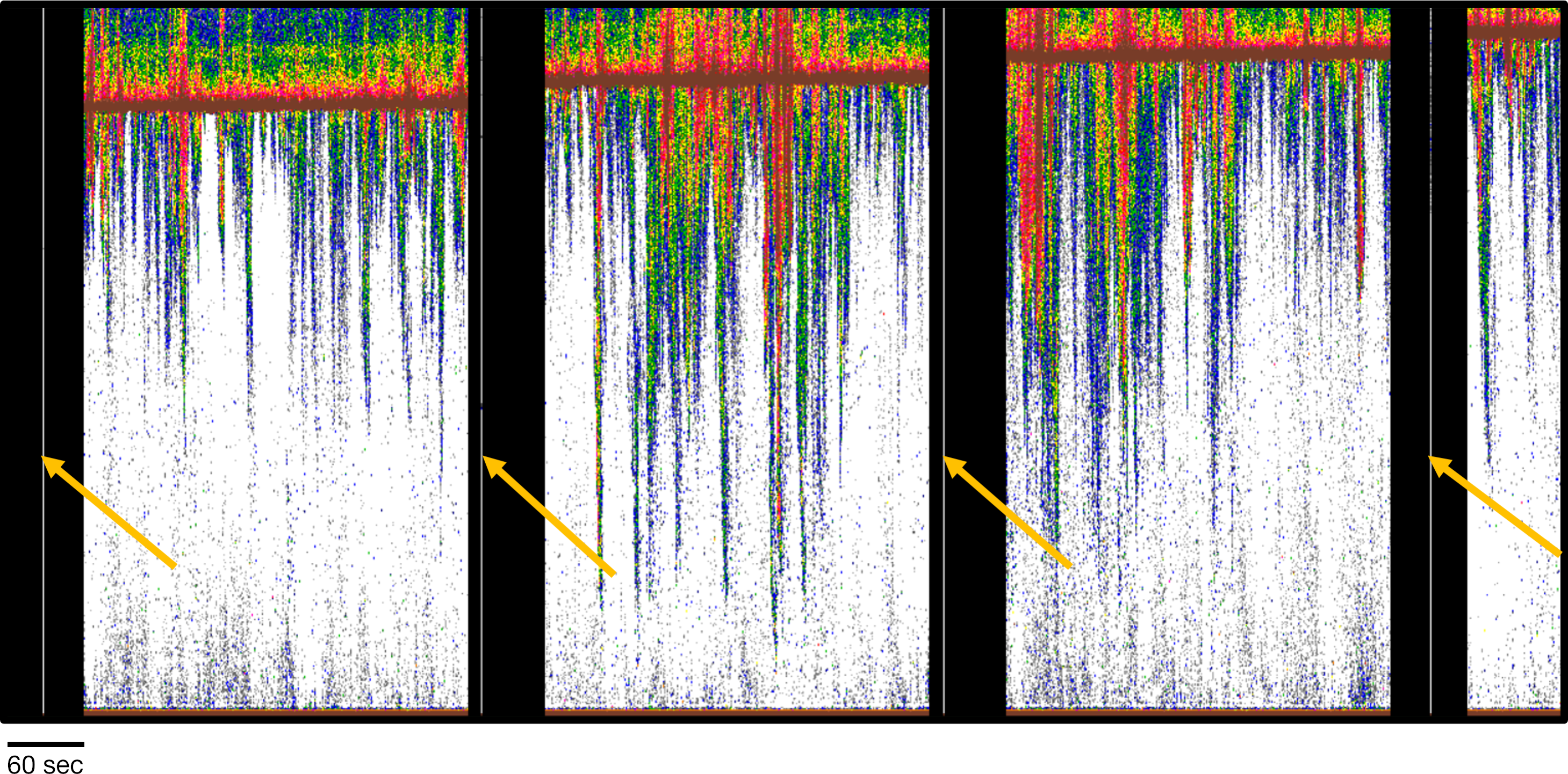}
    \caption{Passive data regions (black vertical bars) as identified by Echoview algorithms. Note the white vertical lines marked by yellow arrows within the black passive data regions. The white vertical lines are single pings or a few pings misclassified by the Echoview algorithm.
    Data: stationary data with echosounder in upfacing orientation, active data recorded for 5 minutes and passive data for one minute every half hour at the Minas Passage site.}
    \label{fig:passiveregions}
\end{figure}

\begin{figure}[!tbp]
    \centering
    \includegraphics[width=\textwidth]{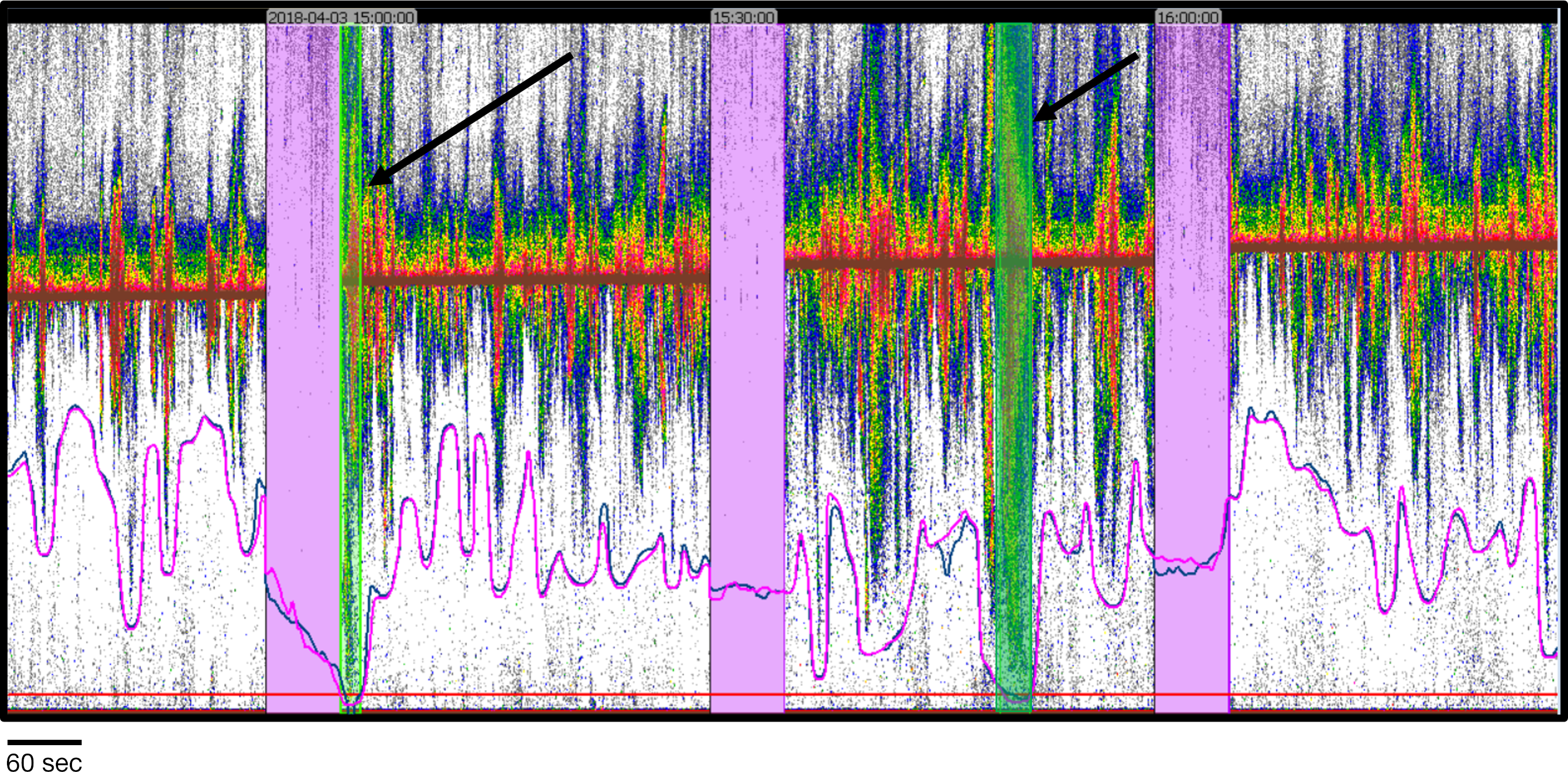}
    \caption{Example of passive data regions (pink bars) and bad data periods (green bars) defined by Echofilter. Echofilter successfully identified the contiguous pings of passive data collection. Bad data periods designated by Echofilter were pings where the entrained-air line (pink or blue lines in this example) intersects with the bottom analytical line (in this case the line designating the outer boundary of the transducer nearfield). Although these cases meet the criteria, in each case the hydroacoustic analyst would designate the entire recording period as a bad data region due to the strength, penetration, and persistence of entrained-air signals.
    Data: stationary data with echosounder in upfacing orientation, active data recorded for 5 minutes and passive data for one minute every half hour at the Minas Passage site.}
    \label{fig:baddatarectangles}
\end{figure}

\begin{figure}[!tbp]
    \centering
    \includegraphics[width=\textwidth]{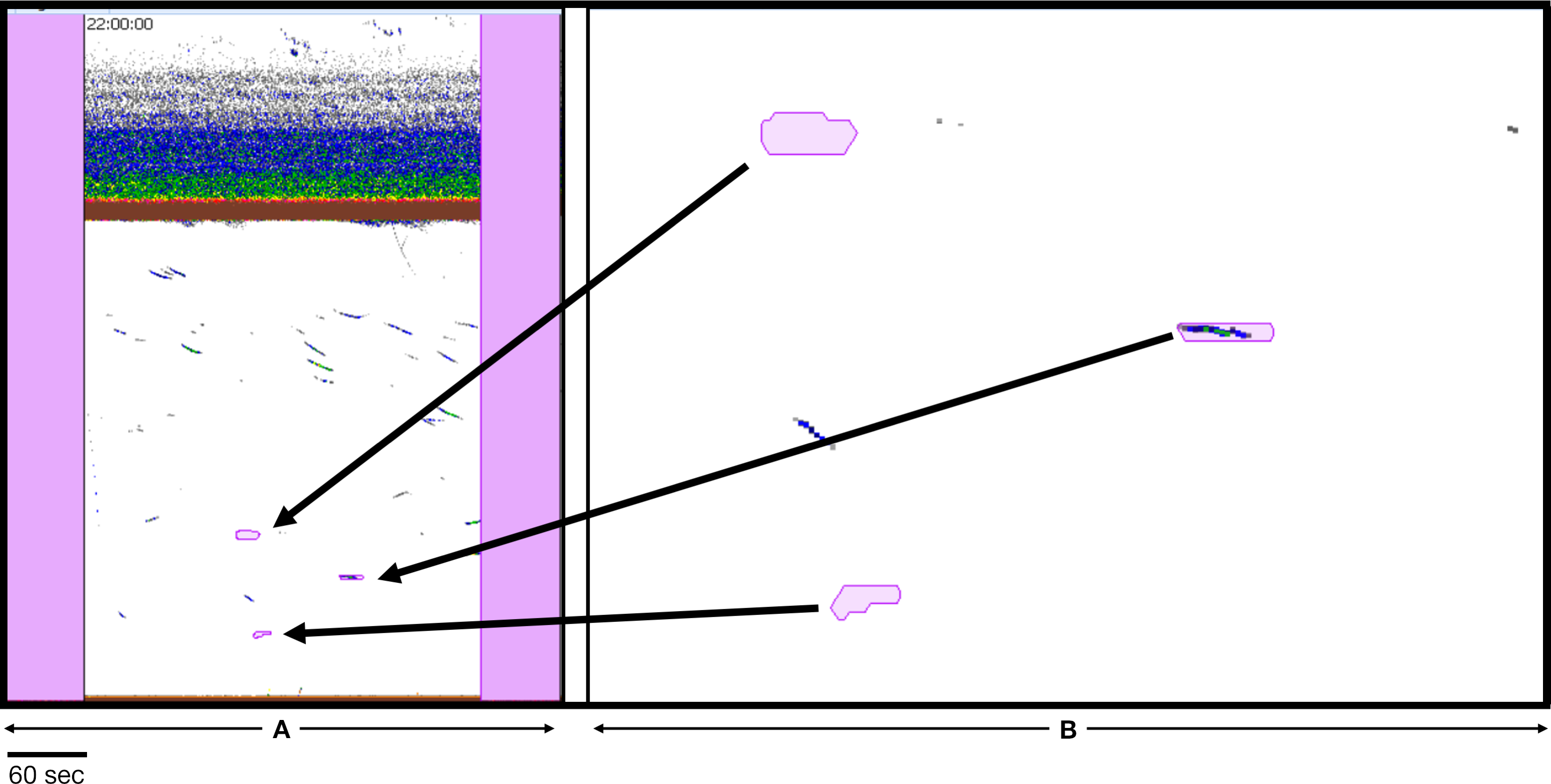}
    \caption{Example of false positive ``patch'' bad data regions identified by Echofilter. \textbf{(A)} A 5-minute section of echogram with passive data regions (pink rectangles) on either side. \textbf{(B)} Enlargement to show the contents within each patch. Empty patches are false positive. The patch containing color samples within it would be classified as fish by the hydroacoustic analyst. It was likely identified as a bad data region by Echofilter because of its nearly horizontal position. The data on which the models were trained contain occurrences of unidentified interference which appear as horizontal lines. Those were classified as bad data regions by the analyst prior to training. Both models (Bifacing@100ep and Upfacing@100ep) designate true and false positives, but differently. Bifacing@100ep results appear to include fewer false positives.
    Data: stationary data with echosounder in upfacing orientation, active data recorded for 5 minutes and passive data for one minute at Minas Passage Site.}
    \label{fig:baddatapatches}
\end{figure}

During model development, a series of iterative testing and upgrades to Echofilter was undertaken. Echofilter was run on the entire set of test files, applying both models (Bifacing and Upfacing, with thresholded zoom+repeat, and logit-smoothing) to the data for comparative purposes.
The results were examined for adequacy and appropriateness of the placement of lines (sea surface and entrained-air), the identification of the passive data collection periods and identification of bad data regions.
Issues with the outputs were investigated in detail, and used to make changes to the model architecture design, training paradigm, or to the format of input and target data provided to the model during training.
This process was iterated until any additional improvements were marginal and inconsequential.

By the end of testing and upgrades to the models, both models (Bifacing and Upfacing) produced appropriate automated initial placement of the boundary lines. Most importantly, the model placement of entrained-air boundary lines were visibly superior to the line placements as produced by the Echoview algorithms, as shown in \autoref{fig:turbulenceline}. The model results proved to be much more responsive than the Echoview algorithms to the entrained-air ambit characteristics across the varying tidal flow rates (\autoref{fig:turbulenceline}). In some cases, the automated prediction of the entrained-air line placement as produced by Echofilter were far superior to that produced by Echoview; see \autoref{fig:echofilterwins}. Note that Echofilter entrained-air lines as defined by each model (Bifacing and Upfacing) were essentially equivalent, although not identical.

As shown in \autoref{fig:surfaceline}, the Echofilter models produced appropriate and adequate automated placement of the surface line, including a user-defined offset; in this case \SI{1}{\metre}. Likewise, the Echofilter models produced appropriate and adequate identification of the passive data regions that will be excluded from biological analyses.  We found the Echofilter passive data region identification was superior to the Echoview algorithms implemented to automate the identification of passive data regions. The Echoview algorithms would, not uncommonly, exclude a ping or few pings from within the passive data region, thereby inappropriately designating those pings for inclusion in biological analyses, as shown in \autoref{fig:passiveregions}. No such occurrences were noted in the Echofilter results (\eg{} \autoref{fig:baddatarectangles}).

In addition to the passive data regions, there are two additional types of bad data regions that are not uncommon to echosounder data. The first type, is a contiguous time period marked to be removed from analysis. As shown in \autoref{fig:baddatarectangles}, these bad data regions are identified by Echofilter when the position of the entrained-air line resolves to a position intersecting or extending below the bottom line, whether that line is the seafloor or the line designating the transducer nearfield exclusion line. In other words, when the position of the entrained-air line indicates that the entrained air has penetrated the entire depth of the water column. Such occurrences are not uncommon in the Minas Passage and Grand Passage datasets, sometimes occurring for just a few pings and other times the penetration occurs throughout an entire 5-minute data collection period. The single criteria of intersecting or penetrating below the bottom line is insufficient for defining all pings that should be excluded in their entirety. \autoref{fig:baddatarectangles} provides an example of just such a case: less than 50\% of the water column remains after the entrained-air exclusion. In that case, if the goal of the analyses is to understand metrics within the full water column, that data collection period would need to be excluded in its entirety. 

The second type of bad data region, a ``patch'' of bad data, can be characterized as forming randomly shaped discrete patches. Within the original test segment of 24 files, only three had occurrences of the patch-type bad data region. Two additional EV files containing patch-type bad data regions were identified from the validation and training segments for manual inspection of the patch-type results only. Both Echofilter models performed poorly, generating false positives as illustrated in \autoref{fig:baddatapatches}.

\subsection{Time-savings analysis}
\label{s:timesaving}

We sought to evaluate the amount of time-savings that the Echofilter model would offer, relative to the existing workflow using Echoview algorithms.
Five of the Echoview files from the \dsmp{} test partition were selected for a time test.
The files were selected to represent each tide and phase combination: flooding spring tide, ebbing spring tide, flooding neap tide, and ebbing neap tide, plus one file with especially noisy data for which neither Echoview or Echofilter would likely render a well-placed entrained-air line. 
Annotations were initialized twice: once using the preexisting workflow utilizing Echoview algorithms\hide{ \autoref{fig:dataflow}}, and once using Echofilter with the Upfacing@100ep model, with logit-smoothing enabled.
The initial entrained-air line in each of the ten files was audited and edited by the hydroacoustic analyst (JD), while recording the amount of time taken to do so.
We randomized the order in which tasks (file and seed annotation source) were completed, except the especially noisy ``bad file'' which was evaluated later.

\begin{table}[tb]
  \centering
  \caption{%
Results from the time-to-edit experiment.
A hydroacoustic analyst used the entrained-air lines produced by either Echoview or Echofilter to seed their annotations.
We compare the amount of time needed to convert the seed lines into ``correct'' annotation lines.
Bold: best model (shortest duration).
}
\label{tab:test-perf}
\centerline{
\begin{tabular}{llrrrrr}
\toprule
       &                  & \multicolumn{2}{c}{Edit order} & \multicolumn{2}{c}{Edit time (MM:SS)} & \\
\cmidrule(r){3-4}
\cmidrule(r){5-6}
Tide   & Phase            & Echoview  & Echofilter& Echoview  & Echofilter& Reduction \\
\midrule
Spring & Flood            & 2         & 6         & 8:06      &\textbf{4:04} & 50\% \\
Spring & Ebb              & 3         & 4         & 8:00      &\textbf{4:04} & 49\% \\
Neap   & Flood            & 8         & 7         & 8:18      &\textbf{4:30} & 46\% \\
Neap   & Ebb              & 1         & 5         & 7:08      &\textbf{3:57} & 45\% \\
\midrule
\textbf{Mean} &           &           &           & 7:53      &   \mbf{4:09} & 47\% \\
\midrule
\multicolumn{2}{l}{Bad file}& 9       & 10        & 4:28      &\textbf{1:51} & 59\% \\
\midrule
\multicolumn{2}{l}{\textbf{Overall Mean}} & &     & 7:12      &   \mbf{3:42} & 49\% \\
\bottomrule
\end{tabular}
}
\end{table}

Our results, shown in \autoref{tab:test-perf}, demonstrate that using the annotations generated by Echofilter results in less time taken for the human annotator to complete their task.
For typical data files, the time taken to finalize annotations was consistently 45\%--50\% shorter when using annotations produced by Echofilter as the seed instead of annotations produced by Echoview.
For an especially noisy file, the reduction in time was even larger, at 59\%.
The reduction in time was statistically significant ($p\!<\!0.001$; paired Student's $t$-test).


\section{Discussion}

\subsection{Impact of Echofilter model}

We have described the implementation of a deep learning model, Echofilter, which can be used to generate annotations to segment entrained air appearing in hydroacoustic recordings at tidal energy sites.
Our goal was to produce an automated, model-based approach to the placement of a line appropriately defining the boundary between that portion of the water column contaminated by acoustic returns from entrained air, and that portion of the water column appropriate for biological analyses.
This was motivated by the need for reliable, timely analyses and subsequent reporting to assist regulators, developers, and stakeholders in understanding the risks to fish imposed by the deployment of tidal energy devices into marine ecosystems.

We found the deep learning models we implemented produced significantly and appreciably better placement of the entrained-air line than the Echoview algorithms.
For mobile, downward-facing recordings, the average error was \SI{0.33}{\meter}, less than a third of Echoview's \SI{1.2}{\meter} average error.
For stationary, upward-facing recordings, the average error was \SIrange{0.5}{1.0}{\meter} depending on dataset, consistently less than half the error seen with Echoview algorithms (\SIrange{1.2}{2.2}{\meter}).
Furthermore, the surface, seafloor, and passive region placement were also superior to those produced using Echoview.
The model’s overall annotations had a high level of agreement with the human segmentation, with an intersection-over-union score of 99\% for mobile downfacing recordings and 92\% to 95\% for stationary upfacing recordings.
As such, Echofilter provides a complete automated line placement and passive data identification methodology.

The most challenging segmentation line to place correctly is the entrained-air line, which currently can require time-consuming manual placement due to the lack of a well-placed automated solution.
We found that the increase in accuracy of the automated placement of the entrained-air line provided by Echofilter corresponded to a 50\% reduction in the time required for a hydroacoustician to audit and correct the line placement.
Our quantitative analysis has shown that the Echofilter models produce lines which are closer (in distance) to the line placement defined by the human expert.
Additionally, we note that the ML models are more sensitive to the fine-scale nuances in the boundary position of the entrained air; when the model places the line incorrectly, the errors tend follow the correct shape of the entrained air but are offset by some amount, and hence require only a simple, coarse edit to shift the line in some region to the correct offset.
In contrast, when the Echoview algorithm is incorrect, the shape is incorrect and corrections to the line involve time-consuming fine-scale edits instead.
Since coarse-scale edits are less cognitively taxing and far fewer edits are required, far less analyst fatigue is invoked during manual corrections of the model-placed entrained-air line, thereby allowing the analyst to bring the full-force of their intellect, training, and analytical skills to modifying placement of the line segments as necessary.
Additionally, the reduction in the number of fine-scale edits provides the opportunity for an increase in the standardization and repeatability of line placement, within an analyst's work and among analysts.

Machine learning applied to the hydroacoustic data by which we quantify fish distribution and abundance has garnered improvements to the work flow and increased the efficiency of the work by 50\%, improvements that haven't been achieved any other way.
The machine learning contribution to assessing the ecological impacts of introducing marine renewable energy devices into the marine habitat is the improved analytical consistency and substantial improvements in the timeliness of analyses and subsequent reporting.

\subsection{Limitations associated with Echofilter}

We developed Echofilter with the goal of increasing the efficiency and standardization of the post-processing of hydroacoustic data collected in dynamic marine environments such as tidal channels.
The model was thoroughly evaluated on data recorded from upward-facing stationary echosounders at two tidal energy demonstration sites in the Bay of Fundy.
The models have not been evaluated on data collected in other regions, with other instrumentation, or in other deployment configurations.
Consequently, the performance of Echofilter on data collected under conditions that differ substantially from those used for model development may be heavily impacted and require some level of re-training to ensure accurate results, which is a non-trivial procedure.

In addition to the entrained-air boundary line, Echofilter predicts the depths of the surface (for upfacing recordings) and the seafloor (downfacing).
Our performance metrics indicate that these lines are all placed accurately, however we have not thoroughly inspected the model's output on downfacing recordings and can not confirm the integrity of the seafloor line.

In addition to the lines, our model attempts to predict regions which should be excluded from biological analyses.
However, it was not possible for the model to learn these annotations with sufficient accuracy to be usable for downstream tasks.
Consequently, it is not possible to automate away a need for manual inspection of the data.
A hydroacoustician must always inspect the recordings themselves in order to annotate regions to exclude from analysis, and adjust lines as necessary.

\FloatBarrier

\subsection{Accessing Echofilter}

To ensure the broader community can utilize our model described in this paper, we have released the final implementation, Echofilter, under the \href{https://www.gnu.org/licenses/agpl-3.0.en.html}{AGPLv3} license.
Python source code and a stand-alone Windows executable are available at \url{https://github.com/DeepSenseCA/echofilter}.
Additionally, the command line interface (CLI) and application programming interface (API) documentation is available at \url{https://DeepSenseCA.github.io/echofilter/}.

We hope this tool will prove useful to tidal energy researchers, and the wider hydroacoustic community.

\section*{Data Availability Statement}

The raw data supporting the conclusions of this article can be found online at \url{https://data.fundyforce.ca/forceCloud/index.php/s/BzC87LpbGtnFsjT}.
Queries regarding dataset access should be directed to FORCE, info@fundyforce.ca.


\section*{Author Contributions}

Manuscript written by SCL and LPM.
Data conversion was performed by JN and SCL.
Model architecture design and training was performed by SCL.
Interface and API development by SCL.
Model evaluation was performed by LPM and JD.
Identification of features of specific value to the hydroacoustics community was performed by LPM.
Project oversight by DJH, CW, and SO.

\section*{Funding}

SCL was funded by Mitacs award IT16140. LPM was supported by the Offshore Energy Research Association of Nova Scotia (OERA) under the Pathway Program; funding provided by Natural Resources Canada, grant number ERPP-RA-05.

\section*{Acknowledgments}

This work was supported in part by Mitacs through the Mitacs Accelerate program, and by Offshore Energy Research Association (OERA; \url{https://oera.ca/}).
Computations were performed on the DeepSense (\url{https://deepsense.ca/}) high-performance computing platform. DeepSense is funded by ACOA, the Province of Nova Scotia, the Centre for Ocean Ventures and Entrepreneurship (COVE), IBM Canada~Ltd., and the Ocean Frontier Institute (OFI).
We thank Echoview Software~Pty~Ltd. and Haley Viehman, Ph.D., for providing support and technical advice for this project.
We thank Paul C. Hines of Hines Ocean S\&T for his helpful discussions during preparation of the manuscript.
Additionally, we thank Tyler Boucher, Aur\'elie Daroux, Mili Sanchez, and Haley Viehman, for (in addition to JD and LPM) creating the expert annotations for the data used in this study.

\section*{Conflict of Interest Statement}

The authors declare that the research was conducted in the absence of any commercial or financial relationships that could be construed as a potential conflict of interest.




\bibliographystyle{frontiersinSCNS_ENG_HUMS} 
\bibliography{main}

\begin{thebibliography}{47}
\providecommand{\natexlab}[1]{#1}
\expandafter\ifx\csname urlstyle\endcsname\relax
  \providecommand{\doi}[1]{doi:\discretionary{}{}{}#1}\else
  \providecommand{\doi}{doi:\discretionary{}{}{}\begingroup
  \urlstyle{rm}\Url}\fi
\providecommand{\selectlanguage}[1]{\relax}
\providecommand{\bibAnnoteFile}[1]{%
  \IfFileExists{#1}{\begin{quotation}\noindent\textsc{Key:} #1\\
  \textsc{Annotation:}\ \input{#1}\end{quotation}}{}}
\providecommand{\bibAnnote}[2]{%
  \begin{quotation}\noindent\textsc{Key:} #1\\
  \textsc{Annotation:}\ #2\end{quotation}}

\bibitem[{Bengio et~al.(2021)Bengio, Lecun, and Hinton}]{turinglecture2021}
Bengio, Y., Lecun, Y., and Hinton, G. (2021).
\newblock Deep learning for {AI}.
\newblock \emph{Communications of the ACM} 64, 58–65.
\newblock \doi{10.1145/3448250}
\bibAnnoteFile{turinglecture2021}

\bibitem[{Benoit-Bird and Lawson(2016)}]{Benoit-Bird}
Benoit-Bird, K.~J. and Lawson, G.~L. (2016).
\newblock Ecological insights from pelagic habitats acquired using active
  acoustics.
\newblock \emph{Annual Review of Marine Science} 8, 463--490.
\newblock \doi{10.1146/annurev-marine-122414-034001}
\bibAnnoteFile{Benoit-Bird}

\bibitem[{Blaber et~al.(2000)Blaber, Cyrus, Albaret, Ching, Day, Elliott
  et~al.}]{Blaber2000}
Blaber, S. J.~M., Cyrus, D.~P., Albaret, J.-J., Ching, C.~V., Day, J.~W.,
  Elliott, M., et~al. (2000).
\newblock {Effects of fishing on the structure and functioning of estuarine and
  nearshore ecosystems}.
\newblock \emph{ICES Journal of Marine Science} 57, 590--602.
\newblock \doi{10.1006/jmsc.2000.0723}
\bibAnnoteFile{Blaber2000}

\bibitem[{Cada et~al.(2007)Cada, Ahlgrimm, Bahleda, Bigford, Stavrakas, Hall
  et~al.}]{Cada2007}
Cada, G., Ahlgrimm, J., Bahleda, M., Bigford, T., Stavrakas, S.~D., Hall, D.,
  et~al. (2007).
\newblock Potential impacts of hydrokinetic and wave energy conversion
  technologies on aquatic environments.
\newblock \emph{Fisheries} 32, 174--181.
\newblock \doi{10.1577/1548-8446(2007)32[174:PIOHAW]2.0.CO;2}
\bibAnnoteFile{Cada2007}

\bibitem[{Copping et~al.(2020)Copping, Hemery, Overhus, Garavelli, Freeman,
  Whiting et~al.}]{Copping2020}
Copping, A.~E., Hemery, L.~G., Overhus, D.~M., Garavelli, L., Freeman, M.~C.,
  Whiting, J.~M., et~al. (2020).
\newblock Potential environmental effects of marine renewable energy
  development—the state of the science.
\newblock \emph{Journal of Marine Science and Engineering} 8.
\newblock \doi{10.3390/jmse8110879}
\bibAnnoteFile{Copping2020}

\bibitem[{Cornett et~al.(2015)Cornett, Toupin, and Nistor}]{Cornett2015}
Cornett, A., Toupin, M., and Nistor, I. (2015).
\newblock Appraisal of the {IEC} technical specification for tidal energy
  resource assessment at {M}inas {P}assage, {B}ay of {F}undy, {C}anada.
\newblock In \emph{Proceedings of 2015 European Wave and Tidal Energy
  Conference (EWTEC), Nantes, France}
\bibAnnoteFile{Cornett2015}

\bibitem[{DFO(2008)}]{DFO2008}
DFO (2008).
\newblock \emph{Potential Impacts of, and Mitigation Strategies for,
  Small-Scale Tidal Generation Projects on Coastal Marine Ecosystems in the Bay
  of Fundy}.
\newblock Tech. rep., Fisheries and Oceans Canada (DFO) Canadian Science
  Advisory Secretariat. Science Response 2008/013
\bibAnnoteFile{DFO2008}

\bibitem[{DFO(2018)}]{DFO2018}
DFO (2018).
\newblock \emph{Delineating Important Ecological Features of the
  {Evangeline-Cape Blomidon-Minas Basin} Ecologically and Biologically
  Significant Area ({EBSA})}.
\newblock Tech. rep., Fisheries and Oceans Canada (DFO) Canadian Science
  Advisory Secretariat. Science Response 2018/005
\bibAnnoteFile{DFO2018}

\bibitem[{Fernandes et~al.(2002)Fernandes, Gerlotto, Holliday, Nakken, and
  Simmonds}]{Fernandes2002}
Fernandes, P., Gerlotto, F., Holliday, D., Nakken, O., and Simmonds, E. (2002).
\newblock Acoustic applications in fisheries science: the {ICES} contribution.
\newblock \emph{ICES Marine Science Symposia} 215, 483--492.
\newblock \doi{10.17895/ices.pub.8889}
\bibAnnoteFile{Fernandes2002}

\bibitem[{Goodfellow et~al.(2016)Goodfellow, Bengio, and
  Courville}]{Goodfellow-book}
Goodfellow, I., Bengio, Y., and Courville, A. (2016).
\newblock \emph{Deep {L}earning} (Cambridge, MA, USA: MIT Press).
\newblock \url{http://deeplearningbook.org}
\bibAnnoteFile{Goodfellow-book}

\bibitem[{Guerra et~al.(2021)Guerra, Hay, Karsten, Trowse, and Cheel}]{Guerra}
Guerra, M., Hay, A.~E., Karsten, R., Trowse, G., and Cheel, R.~A. (2021).
\newblock Turbulent flow mapping in a high-flow tidal channel using mobile
  acoustic doppler current profilers.
\newblock \emph{Renewable Energy} 177, 759--772.
\newblock \doi{10.1016/j.renene.2021.05.133}
\bibAnnoteFile{Guerra}

\bibitem[{He et~al.(2016)He, Zhang, Ren, and Sun}]{resnet}
He, K., Zhang, X., Ren, S., and Sun, J. (2016).
\newblock Deep residual learning for image recognition.
\newblock In \emph{Proceedings of the IEEE Conference on Computer Vision and
  Pattern Recognition (CVPR)} (New York, NY, USA: IEEE), 770--778.
\newblock \doi{10.1109/CVPR.2016.90}
\bibAnnoteFile{resnet}

\bibitem[{Howard et~al.(2017)Howard, Zhu, Chen, Kalenichenko, Wang, Weyand
  et~al.}]{mobilenet}
Howard, A.~G., Zhu, M., Chen, B., Kalenichenko, D., Wang, W., Weyand, T.,
  et~al. (2017).
\newblock {MobileNets}: Efficient convolutional neural networks for mobile
  vision applications.
\newblock \emph{arXiv preprint} arXiv:1704.04861.
\newblock \doi{10.48550/arxiv.1704.04861}
\bibAnnoteFile{mobilenet}

\bibitem[{Hu et~al.(2019)Hu, Shen, Albanie, Sun, and Wu}]{squeezeandexcitation}
Hu, J., Shen, L., Albanie, S., Sun, G., and Wu, E. (2019).
\newblock Squeeze-and-excitation networks.
\newblock \emph{arXiv preprint} arXiv:1709.01507.
\newblock \doi{10.48550/arxiv.1709.01507}
\bibAnnoteFile{squeezeandexcitation}

\bibitem[{{IPCC}(2021)}]{ipcc2021}
{IPCC} (2021).
\newblock Summary for policymakers.
\newblock In \emph{Climate Change 2021: The Physical Science Basis.
  Contributions of Working Group {I} to the Sixth Assessment Report of the
  {I}ntergovernmental {P}anel on {C}limate {C}hange}, eds. V.~Masson-Delmotte,
  P.~Zhai, A.~Pirani, S.~L. Connors, C.~Péan, S.~Berger, N.~Caud, Y.~Chen,
  L.~Goldfarb, M.~I. Gomis, M.~Huang, K.~Leitzell, E.~Lonnoy, J.~B.~R.
  Matthews, T.~K. Maycock, T.~Waterfield, O.~Yelekçi, R.~Yu, and B.~Zhou
  (Cambridge, United Kingdom and New York, NY, USA: Cambridge University
  Press). 3--32
\bibAnnoteFile{ipcc2021}

\bibitem[{IRENA(2020)}]{IRENA2020}
IRENA (2020).
\newblock \emph{Innovation outlook: Ocean energy technologies}.
\newblock Tech. rep., International Renewable Energy Agency (IRENA), Abu Dhabi
\bibAnnoteFile{IRENA2020}

\bibitem[{Jaccard(1912)}]{jaccard}
Jaccard, P. (1912).
\newblock The distribution of the flora in the alpine zone.
\newblock \emph{New Phytologist} 11, 37--50.
\newblock \doi{10.1111/j.1469-8137.1912.tb05611.x}
\bibAnnoteFile{jaccard}

\bibitem[{Johannesson and Mitson(1983)}]{Johannesson1983}
Johannesson, K.~A. and Mitson, R.~B. (1983).
\newblock \emph{Fisheries Acoustics: A practical manual for aquatic biomass
  estimation}, vol. 240 of \emph{FAO Fisheries Technical Paper} (Rome, Italy:
  Food and Agriculture Organization of the United Nations)
\bibAnnoteFile{Johannesson1983}

\bibitem[{Karpathy(2014)}]{KarpathyBlog2014}
Karpathy, A. (2014).
\newblock What {I} learned from competing against a {ConvNet} on {ImageNet}.
\newblock \emph{Andrej Karpathy blog}
  \url{http://karpathy.github.io/2014/09/02/what-i-learned-from-competing-against-a-convnet-on-imagenet/},
  Accessed: 2022-05-01
\bibAnnoteFile{KarpathyBlog2014}

\bibitem[{Karsten et~al.(2013)Karsten, Swan, and Culina}]{Karsten2013}
Karsten, R., Swan, A., and Culina, J. (2013).
\newblock Assessment of arrays of in-stream tidal turbines in the bay of fundy.
\newblock \emph{Philosophical Transactions of the Royal Society A:
  Mathematical, Physical and Engineering Sciences} 371, 20120189.
\newblock \doi{10.1098/rsta.2012.0189}
\bibAnnoteFile{Karsten2013}

\bibitem[{Krizhevsky et~al.(2012)Krizhevsky, Sutskever, and Hinton}]{alexnet}
Krizhevsky, A., Sutskever, I., and Hinton, G.~E. (2012).
\newblock Image{N}et classification with deep convolutional neural networks.
\newblock In \emph{Advances in Neural Information Processing Systems}, eds.
  F.~Pereira, C.~Burges, L.~Bottou, and K.~Weinberger (Red Hook, NY, USA:
  Curran Associates, Inc.), vol.~25
\bibAnnoteFile{alexnet}

\bibitem[{LeCun et~al.(2015)LeCun, Bengio, and Hinton}]{deeplearning2015}
LeCun, Y., Bengio, Y., and Hinton, G. (2015).
\newblock Deep learning.
\newblock \emph{Nature} 521, 436--444.
\newblock \doi{10.1038/nature14539}
\bibAnnoteFile{deeplearning2015}

\bibitem[{Liu et~al.(2020)Liu, Jiang, He, Chen, Liu, Gao et~al.}]{RAdam}
Liu, L., Jiang, H., He, P., Chen, W., Liu, X., Gao, J., et~al. (2020).
\newblock On the variance of the adaptive learning rate and beyond.
\newblock In \emph{8th International Conference on Learning Representations,
  {ICLR} 2020, Addis Ababa, Ethiopia, April 26-30, 2020} (OpenReview).
\newblock \doi{10.48550/arXiv.1908.03265}
\bibAnnoteFile{RAdam}

\bibitem[{Lowe et~al.(2021)Lowe, Trappenberg, and Oore}]{logavgexp}
Lowe, S.~C., Trappenberg, T., and Oore, S. (2021).
\newblock {LogAvgExp} provides a principled and performant global pooling
  operator.
\newblock \emph{arXiv preprint} arXiv:2111.01742.
\newblock \doi{10.48550/arxiv.2111.01742}
\bibAnnoteFile{logavgexp}

\bibitem[{Melvin and Cochrane(2012)}]{Melvin2012}
Melvin, G.~D. and Cochrane, N.~A. (2012).
\newblock \emph{A Preliminary Investigation of Fish Distributions Near an
  In-Stream Tidal Turbine in {Minas Passage, Bay of Fundy}}.
\newblock Tech. rep., Canadian Technical Report of Fisheries and Aquatic
  Sciences 3006
\bibAnnoteFile{Melvin2012}

\bibitem[{Melvin and Cochrane(2015)}]{Melvin2015}
Melvin, G.~D. and Cochrane, N.~A. (2015).
\newblock Multibeam acoustic detection of fish and water column targets at
  high-flow sites.
\newblock \emph{Estuaries and Coasts} 38, 227--240.
\newblock \doi{10.1007/s12237-014-9828-z}
\bibAnnoteFile{Melvin2015}

\bibitem[{Minaee et~al.(2022)Minaee, Boykov, Porikli, Plaza, Kehtarnavaz, and
  Terzopoulos}]{deep-segmentation-survey}
Minaee, S., Boykov, Y., Porikli, F., Plaza, A., Kehtarnavaz, N., and
  Terzopoulos, D. (2022).
\newblock Image segmentation using deep learning: A survey.
\newblock \emph{IEEE Transactions on Pattern Analysis and Machine Intelligence}
  44, 3523--3542.
\newblock \doi{10.1109/TPAMI.2021.3059968}
\bibAnnoteFile{deep-segmentation-survey}

\bibitem[{Perez et~al.(2021)Perez, Cossu, Grinham, and Penesis}]{Perez2021}
Perez, L., Cossu, R., Grinham, A., and Penesis, I. (2021).
\newblock Seasonality of turbulence characteristics and wave-current
  interaction in two prospective tidal energy sites.
\newblock \emph{Renewable Energy} 178, 1322--1336.
\newblock \doi{10.1016/j.renene.2021.06.116}
\bibAnnoteFile{Perez2021}

\bibitem[{Redmon et~al.(2016)Redmon, Divvala, Girshick, and Farhadi}]{yolo}
Redmon, J., Divvala, S., Girshick, R., and Farhadi, A. (2016).
\newblock You only look once: Unified, real-time object detection.
\newblock In \emph{Proceedings of the IEEE Conference on Computer Vision and
  Pattern Recognition (CVPR)}. 779--788.
\newblock \doi{10.1109/CVPR.2016.91}
\bibAnnoteFile{yolo}

\bibitem[{Roberts et~al.(2016)Roberts, Thomas, Sewell, Khan, Balmain, and
  Gillman}]{Roberts2016}
Roberts, A., Thomas, B., Sewell, P., Khan, Z., Balmain, S., and Gillman, J.
  (2016).
\newblock Current tidal power technologies and their suitability for
  applications in coastal and marine areas.
\newblock \emph{Journal of Ocean Engineering and Marine Energy} 2, 227--245.
\newblock \doi{10.1007/s40722-016-0044-8}
\bibAnnoteFile{Roberts2016}

\bibitem[{Ronneberger et~al.(2015)Ronneberger, Fischer, and Brox}]{unet}
Ronneberger, O., Fischer, P., and Brox, T. (2015).
\newblock U-{N}et: Convolutional networks for biomedical image segmentation.
\newblock In \emph{Medical Image Computing and Computer-Assisted Intervention
  -- MICCAI 2015}, eds. N.~Navab, J.~Hornegger, W.~M. Wells, and A.~F. Frangi
  (Cham: Springer International Publishing), 234--241.
\newblock \doi{10.1007/978-3-319-24574-4_28}
\bibAnnoteFile{unet}

\bibitem[{Russakovsky et~al.(2015)Russakovsky, Deng, Su, Krause, Satheesh, Ma
  et~al.}]{ImageNet2014}
Russakovsky, O., Deng, J., Su, H., Krause, J., Satheesh, S., Ma, S., et~al.
  (2015).
\newblock Image{N}et large scale visual recognition challenge.
\newblock \emph{International Journal of Computer Vision} 115, 211--252.
\newblock \doi{10.1007/s11263-015-0816-y}
\bibAnnoteFile{ImageNet2014}

\bibitem[{Santoro et~al.(2016)Santoro, Bartunov, Botvinick, Wierstra, and
  Lillicrap}]{Santoro2016}
Santoro, A., Bartunov, S., Botvinick, M., Wierstra, D., and Lillicrap, T.
  (2016).
\newblock One-shot learning with memory-augmented neural networks.
\newblock \emph{arXiv preprint} arXiv:1605.06065.
\newblock \doi{10.48550/arxiv.1605.06065}
\bibAnnoteFile{Santoro2016}

\bibitem[{Schmidhuber(2015)}]{Schmidhuber2015}
Schmidhuber, J. (2015).
\newblock {D}eep {L}earning.
\newblock \emph{Scholarpedia} 10, 32832.
\newblock \doi{10.4249/scholarpedia.32832}.
\newblock Revision \#184887
\bibAnnoteFile{Schmidhuber2015}

\bibitem[{Silver et~al.(2017)Silver, Schrittwieser, Simonyan, Antonoglou,
  Huang, Guez et~al.}]{AlphaZero}
Silver, D., Schrittwieser, J., Simonyan, K., Antonoglou, I., Huang, A., Guez,
  A., et~al. (2017).
\newblock Mastering the game of {Go} without human knowledge.
\newblock \emph{Nature} 550, 354--359.
\newblock \doi{10.1038/nature24270}
\bibAnnoteFile{AlphaZero}

\bibitem[{Simmonds and MacLennan(2005)}]{Simmonds}
Simmonds, J. and MacLennan, D. (2005).
\newblock \emph{Fisheries Acoustics Theory and Practice} (Oxford, UK: Blackwell
  Publishing), 2nd edn.
\bibAnnoteFile{Simmonds}

\bibitem[{Smith(2015)}]{clr}
Smith, L.~N. (2015).
\newblock No more pesky learning rate guessing games.
\newblock \emph{arXiv preprint} arXiv:1506.01186.
\newblock \doi{10.48550/arxiv.1506.01186}
\bibAnnoteFile{clr}

\bibitem[{Smith(2018)}]{onecycle}
Smith, L.~N. (2018).
\newblock A disciplined approach to neural network hyper-parameters: Part 1 --
  learning rate, batch size, momentum, and weight decay.
\newblock \emph{arXiv preprint} arXiv:1803.09820.
\newblock \doi{10.48550/arxiv.1803.09820}
\bibAnnoteFile{onecycle}

\bibitem[{Smith and Topin(2017)}]{superconvergence}
Smith, L.~N. and Topin, N. (2017).
\newblock Super-convergence: Very fast training of residual networks using
  large learning rates.
\newblock \emph{arXiv preprint} arXiv:1708.07120.
\newblock \doi{10.48550/arxiv.1708.07120}
\bibAnnoteFile{superconvergence}

\bibitem[{Tan and Le(2019)}]{efficientnet}
Tan, M. and Le, Q. (2019).
\newblock {E}fficient{N}et: Rethinking model scaling for convolutional neural
  networks.
\newblock In \emph{Proceedings of the 36th International Conference on Machine
  Learning (ICML)}, eds. K.~Chaudhuri and R.~Salakhutdinov (PMLR), vol.~97 of
  \emph{Proceedings of Machine Learning Research}, 6105--6114.
\newblock \doi{10.48550/arxiv.1905.11946}
\bibAnnoteFile{efficientnet}

\bibitem[{Tong et~al.(2022)Tong, Liang, and Bi}]{calibratingAGM}
Tong, Q., Liang, G., and Bi, J. (2022).
\newblock Calibrating the adaptive learning rate to improve convergence of
  {ADAM}.
\newblock \emph{Neurocomputing} 481, 333--356.
\newblock \doi{10.1016/j.neucom.2022.01.014}
\bibAnnoteFile{calibratingAGM}

\bibitem[{Tsitrin et~al.(2022)Tsitrin, Sanderson, McLean, Gibson, Hardie, and
  Stokesbury}]{Tsitrin2022}
Tsitrin, E., Sanderson, B.~G., McLean, M.~F., Gibson, A. J.~F., Hardie, D.~C.,
  and Stokesbury, M. J.~W. (2022).
\newblock Migration and apparent survival of post-spawning alewife
  (\textit{Alosa pseudoharengus}) in {Minas Basin, Bay of Fundy}.
\newblock \emph{Animal Biotelemetry} 10:11.
\newblock \doi{10.1186/s40317-022-00277-z}
\bibAnnoteFile{Tsitrin2022}

\bibitem[{Williamson et~al.(2017)Williamson, Fraser, Blondel, Bell, Waggitt,
  and Scott}]{Williamson2017}
Williamson, B.~J., Fraser, S., Blondel, P., Bell, P.~S., Waggitt, J.~J., and
  Scott, B.~E. (2017).
\newblock Multisensor acoustic tracking of fish and seabird behavior around
  tidal turbine structures in {S}cotland.
\newblock \emph{IEEE Journal of Oceanic Engineering} 42, 948--965.
\newblock \doi{10.1109/JOE.2016.2637179}
\bibAnnoteFile{Williamson2017}

\bibitem[{Wolf et~al.(2022)Wolf, Dominicis, Lewis, Neill, {O’Hara Murray},
  Scott et~al.}]{Wolf2022}
Wolf, J., Dominicis, M.~D., Lewis, M., Neill, S.~P., {O’Hara Murray}, R.,
  Scott, B., et~al. (2022).
\newblock 9.04 - {E}nvironmental issues for offshore renewable energy.
\newblock In \emph{Comprehensive Renewable Energy}, ed. T.~M. Letcher (Oxford:
  Elsevier). 2nd edn., 25--59.
\newblock \doi{10.1016/B978-0-12-819727-1.00036-4}
\bibAnnoteFile{Wolf2022}

\bibitem[{Wright(2019)}]{Ranger}
Wright, L. (2019).
\newblock Ranger - a synergistic optimizer.
\newblock \emph{GitHub repository}
  \url{https://github.com/lessw2020/Ranger-Deep-Learning-Optimizer}.
\newblock Revision @8d636a5
\bibAnnoteFile{Ranger}

\bibitem[{Yong et~al.(2020)Yong, Huang, Hua, and Zhang}]{GradientCentral}
Yong, H., Huang, J., Hua, X., and Zhang, L. (2020).
\newblock Gradient centralization: A new optimization technique for deep neural
  networks.
\newblock In \emph{Computer Vision -- ECCV 2020}, eds. A.~Vedaldi, H.~Bischof,
  T.~Brox, and J.-M. Frahm (Cham: Springer International Publishing), 635--652.
\newblock \doi{10.48550/arxiv.2004.01461}
\bibAnnoteFile{GradientCentral}

\bibitem[{Zhang et~al.(2019)Zhang, Lucas, Ba, and Hinton}]{lookahead}
Zhang, M., Lucas, J., Ba, J., and Hinton, G.~E. (2019).
\newblock Lookahead optimizer: k steps forward, 1 step back.
\newblock In \emph{Advances in Neural Information Processing Systems}, eds.
  H.~Wallach, H.~Larochelle, A.~Beygelzimer, F.~d\textquotesingle
  Alch\'{e}-Buc, E.~Fox, and R.~Garnett (Red Hook, NY, USA: Curran Associates,
  Inc.), vol.~32.
\newblock \doi{10.48550/arxiv.1907.08610}
\bibAnnoteFile{lookahead}

\end{thebibliography}


\end{document}